\documentclass[10pt,twocolumn,letterpaper]{article}

\usepackage{cvpr}              

\usepackage{graphicx}
\usepackage{amsmath,amsthm}
\usepackage{amssymb}
\usepackage{booktabs}

\usepackage{bm}
\usepackage{algorithm}
\usepackage{algpseudocode}
\usepackage{xcolor}
\usepackage{multirow}

\definecolor{ForestGreen}{RGB}{34,139,34}
\newcommand{\BNReLUConv}{BN$\rightarrow$ReLU$\rightarrow$Conv}

\newtheorem{corollary}{Corollary}

\newtheorem{proposition}{Proposition}
\newtheorem{definition}{Definition}
\newcommand{\nnMass}{m}

\usepackage[pagebackref,breaklinks,colorlinks]{hyperref}

\usepackage[capitalize]{cleveref}
\crefname{section}{Sec.}{Secs.}
\Crefname{section}{Section}{Sections}
\Crefname{table}{Table}{Tables}
\crefname{table}{Tab.}{Tabs.}

\def\cvprPaperID{*****} 
\def\confName{CVPR}
\def\confYear{2022}

\begin{document}

\title{Restructurable Activation Networks}

\author{Kartikeya Bhardwaj$^1$, James Ward$^1$, Caleb Tung$^{2,}$\thanks{Equal Contribution (alphabetical order). $^{\dag}$Work done while at Arm.}$^{\ ,\dag}$, Dibakar Gope$^{1,*}$, Lingchuan Meng$^{3,\dag}$,\\Igor Fedorov$^{4,\dag}$, Alex Chalfin$^1$, Paul Whatmough$^1$, and Danny Loh$^1$\\
$^1$Arm Inc., $^2$Purdue University,  $^3$Amazon, $^4$Meta\\
{\tt\small kartikeya.bhardwaj@arm.com, dibakar.gope@arm.com}
}
\maketitle

\begin{abstract}
   Is it possible to restructure the non-linear activation functions in a deep network to create hardware-efficient models? To address this question, we propose a new paradigm called Restructurable Activation Networks (RANs) that manipulate the amount of non-linearity in models to improve their hardware-awareness and efficiency. First, we propose RAN-explicit (RAN-e) -- a new hardware-aware search space and a semi-automatic search algorithm -- to replace inefficient blocks with hardware-aware blocks. Next, we propose a training-free model scaling method called RAN-implicit (RAN-i) where we theoretically prove the link between network topology and its expressivity in terms of number of non-linear units. We demonstrate that our networks achieve state-of-the-art results on ImageNet at different scales and for several types of hardware. For example, compared to EfficientNet-Lite-B0, RAN-e achieves a similar accuracy while  improving Frames-Per-Second (FPS) by $1.5\times$ on Arm micro-NPUs. On the other hand, RAN-i demonstrates up to $2\times$ reduction in \textit{\#}MACs  over ConvNexts with a similar or better accuracy. We also show that RAN-i achieves nearly $40\%$ higher FPS than ConvNext on Arm-based datacenter CPUs. Finally, RAN-i based object detection networks achieve a similar or higher mAP and up to $33\%$ higher FPS on datacenter CPUs compared to ConvNext based models\footnote{The code to train and evaluate RANs and the pretrained networks are available at \href{https://github.com/ARM-software/ML-restructurable-activation-networks}{this Github repository}.}. 
\end{abstract}

\section{Introduction}
\label{sec:intro}
Tremendous progress has been made towards building efficient deep networks using either model compression~\cite{pruning, quantization, hinton2015distilling}, manual model design~\cite{mobilenetv2, convnext}, or automatic Neural Architecture Search (NAS)-based techniques~\cite{efficientnet, donna}. Despite these advances, significant challenges remain in (1)~hardware-aware model design especially for AI accelerators like Neural Processing Units (NPUs)~\cite{u55,u65}, and (2)~cost of finding optimized models for a given \textit{\#}MACs/\textit{\#}parameters constraint once a good base model is known. We discuss both of these challenges in detail below.

The first challenge relates to the lack of good hardware-aware building blocks. Specifically, even though excellent NAS methods~\cite{proxylessnas, micronets_mlsys2021, EfficientNetV2ICML2021,FastModelFamiliesCVPR2021, mbdets, fbnetv2, UNASCVPR2020, MixedPrecisionDNNsICLR2020, MCUNetNeurIPS2020, efficientnet, donna} exist to find highly efficient deep networks within large search spaces, there has been limited focus on building a \textit{hardware-aware search space} itself, particularly for AI accelerators. Most search spaces for computer vision tasks rely on Inverted Bottleneck (IBN) blocks -- the main building blocks used in MobileNet-V2~\cite{mobilenetv2} and EfficientNet~\cite{efficientnet} -- since they result in highly accurate yet compact models. Existing NAS works search over number of channels, expansion ratio, and kernel sizes for IBN blocks~\cite{efficientnet, proxylessnas}. However, it has been established that while IBN blocks are great for generic processors like phone CPUs, they are not always well-suited for AI accelerators due to poor utilization of the accelerator hardware~\cite{mbdets,mlsys22Codesign}. To address this, recent NAS techniques use fused convolutions~\cite{mbdets, mlsys22Codesign} which combine the first two layers of the IBN to form a large, regular convolution. This leads to layers that do not present hardware utilization issues but are computationally very expensive in terms of \textit{\#}MACs/\textit{\#}parameters. Hence, a new search space is needed which contains blocks that (a)~enable hardware-aware (i.e., high hardware utilization) models for AI accelerators with low \textit{\#}MACs while achieving high accuracy, and (b)~are accompanied by a simple search algorithm.

The second challenge relates to the fact that even if a good model has been designed (either using NAS or manually), it is still very costly to scale it up or down to satisfy various \textit{\#}MACs/\textit{\#}parameters constraints. For example, some existing works perform model scaling using ad-hoc methods, which can result in suboptimal networks (e.g., in ConvNexts~\cite{convnext}, the number of blocks is scaled from [3,3,9,3] for ConvNext-Tiny to [3,3,27,3] for ConvNext-Small model without an explanation). Other methods rely on extremely costly EfficientNet-like NAS~\cite{efficientnet} to find optimal width and depth scaling. Therefore, more focus is required on inexpensive model scaling techniques that result in high accuracy.

In this paper, we propose a new paradigm to create hardware-efficient deep networks. Specifically, we show that manipulating the amount of non-linearity in deep networks can be a new way to achieve hardware-awareness and/or significant reduction in computational cost. Non-linear activation functions have been a fundamental part of deep neural networks since their inception. While many advanced activation functions have been proposed in literature~\cite{ramachandran2017searching,hendrycks2016bridging,mbv3}, several important questions remain unaddressed. For instance, how much non-linearity can we remove from a network without significant accuracy loss? Activation functions have been viewed as cheap operations in deep learning from a computational cost standpoint. Consequently, they have not been used to build efficient deep networks in prior work. In view of the above challenges, we ask the following \textbf{key questions} in this paper: 
\begin{enumerate}
    \item Is it possible to manipulate the non-linearities in a deep network to create accelerator-hardware-aware models?
    \item Given a good base model, can changing the amount of non-linearity in a network allow us to quickly scale it up or down to any target resource constraints in a \textit{training-free} way to obtain highly accurate models?
\end{enumerate}

To address the above questions, we propose \textit{Restructurable Activation Networks} (RANs). Our approach is based on \textit{explicit} and \textit{implicit} restructuring of the non-linear activation functions to improve the model efficiency while still achieving high accuracy. Specifically, for the first question, we propose a search space that contains new building blocks that can \textit{restructure} IBN blocks into small, regular convolutions to generate hardware-aware networks. This is highly useful to improve the hardware utilization (e.g., for AI accelerators) without increasing \textit{\#}MACs/\textit{\#}parameters significantly. Since the amount of non-linearity and structure of the model explicitly changes with this technique, we call these models RAN-e (RAN-explicit). 

For the second question, we look into a recent study that theoretically analyzes the topological properties of deep networks and shows how accuracy of different models is related to their structural characteristics (e.g., presence of skip connections, etc.)~\cite{nnmass}. Since no training is required to evaluate these topological properties, we use this method to scale a given base model in a \textit{training-free} way. We also show that for a certain class of networks, such topological properties are related to the total number of non-linear units in a network. Therefore, changing the topological structure of networks also impacts the amount of non-linearity and, thus, affects the expressivity of deep networks. Hence, we exploit the metrics in~\cite{nnmass} to scale ConvNext class of models and show that they can be scaled in a significantly better way than the ad-hoc method used in ConvNext~\cite{convnext}. Since our method results in an implicit restructuring of non-linearity, we call these models RAN-i (RAN-implicit).

We emphasize that our work is \textit{not} a full-blown NAS. The objective of this paper is to (a)~demonstrate the power of manipulating the amount of non-linearity in networks to create hardware-efficient models, and (b)~highlight the effectiveness of a new search space that comes with its own lightweight search algorithm towards building accelerator hardware-aware networks. As such, unlike the majority of the NAS literature, we do \textit{not} focus on building the most effective search algorithm. That is, our lightweight search technique is limited only to the proposed blocks within our new search space, is often semi- (and not fully-) automatic, and does \textit{not} search over important factors like number of channels, number of blocks, kernel sizes, expansion ratios, etc. Nevertheless, we demonstrate that our search space and preliminary algorithm result in highly accurate models that perform extremely well on NPUs. Hence, the scope of this work is to design only the new search space that can result in more hardware-aware models. Integrating this search space into a full-blown NAS is left as a future work.

\begin{figure*}[tb]
\centering
\includegraphics[width=0.9\textwidth]{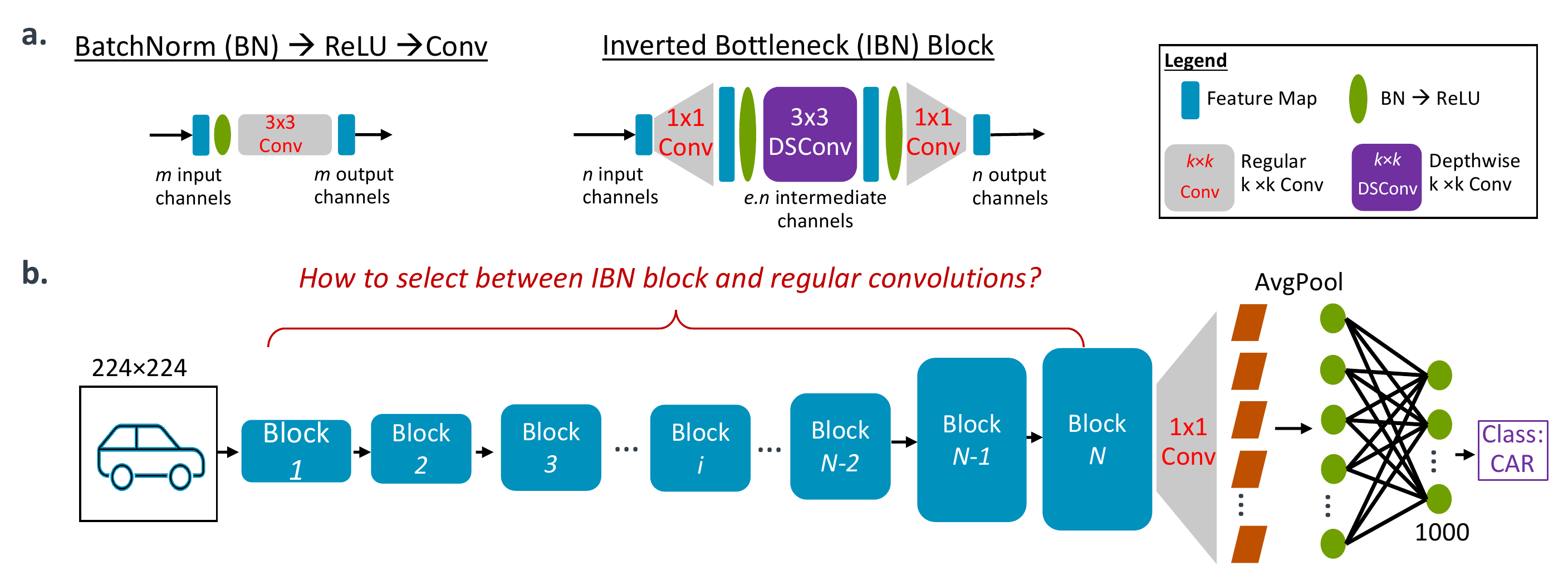}\vspace{-2mm}
	\caption{(a)~Two blocks: BatchNorm (BN)$\rightarrow$ReLU$\rightarrow$Conv and a standard Inverted Bottleneck (IBN) block. Even if the \textit{\#}MACs and \textit{\#}parameters are similar between these two blocks, the IBN block has many more non-linear units than \BNReLUConv. Indeed, the IBN block is much more expressive than a regular convolution layer which is also clear from the high accuracy achieved by the IBN-based models. (b)~We formulate the problem of how much non-linearity we can remove from a network as a search between the blocks in (a).}
\label{fig:setup}
\end{figure*}

We make the following \textbf{key contributions} in this work: 
\begin{enumerate}
    \item We propose Restructurable Activation Networks (RANs), a new paradigm that improves hardware efficiency of deep networks by manipulating the amount of non-linearity in the models. 
    
    \item We first create RAN-explicit (RAN-e) models that rely on a new search space and result in high accuracy and significantly improved accelerator hardware utilization without increasing MACs. We then create RAN-implicit (RAN-i) models that scale existing base models like ConvNexts in a training-free way to satisfy certain \textit{\#}MACs/\textit{\#}parameters. We also present an initial attempt at co-designing a restructurable block with its own activation function for ConvNext networks.
    
    \item While RAN-e results in explicit changes in model structure with direct non-linear unit manipulation, RAN-i modifies the topology (depth/width) of the base model and, hence, implicitly also changes the amount of non-linearity. Towards this implicit non-linearity manipulation, we theoretically prove the link between the topological metric in~\cite{nnmass} and expressivity of deep networks like ResNets and ConvNexts.
    
    \item Finally, we achieve state-of-the-art results on ImageNet at several \textit{\#}MACs/\textit{\#}parameters scales and for multiple types of hardware ranging from micro-NPUs to datacenter CPUs. RAN-e leads to $1.5\times$ higher FPS than EfficientNet-Lite-B0 on an Arm micro-NPU with a similar accuracy. Also, RAN-i outperform ConvNexts by nearly $2\times$ fewer MACs with a minor drop in accuracy ($\sim0.2\%$). We also achieve up to $40\%$ higher FPS than ConvNext on an Arm-based datacenter CPU. When used as backbones in object detection, RAN-i achieve a similar or higher mAP with $33\%$ higher FPS on datacenter CPUs compared to ConvNexts.
\end{enumerate}

The paper is organized as follows. The RAN-e models are proposed in Section~\ref{sec:rane} along with their results. Section~\ref{sec:rani} proposes the RAN-i model scaling and shows its effectiveness. Section~\ref{sec:expCon} demonstrates an initial attempt towards co-designing the restructurable blocks with a new activation function. After some discussion on future directions in Section~\ref{sec:disc}, we review the related work in Section~\ref{sec:rel}. Finally, we conclude the paper in Section~\ref{sec:conc}.

\section{RAN-explicit and New Search Space}
\label{sec:rane}
In this section, we propose RAN-explicit, a new class of models whose architecture can be restructured by manipulating the amount of non-linearity in the network. We accomplish this by proposing a search space that contains new blocks. We start by formulating the problem below.

\subsection{Problem Formulation}
\label{sec:ranePF}
How much non-linearity can we remove from a deep network without losing significant accuracy? To address this question, we first create a setup that can allow us to systematically experiment with the amount of non-linearity in a given model. To this end,  consider the two standard blocks shown in Fig.~\ref{fig:setup}(a): \BNReLUConv\ and IBN. Assuming both blocks receive a feature map of same height and width ($H\times W$), the number of MACs for \BNReLUConv\ = $H\times W\times 3\times 3\times m\times m$ = $9m^2HW$. For the IBN blocks, ignoring the MACs in depthwise layers\footnote{Depthwise layer has far fewer MACs than pointwise $1\times 1$ layers.} and assuming the expansion ratio $e=6$ (similar to~\cite{mobilenetv2,efficientnet}), the number of MACs for IBN = $H\times W\times [1\times 1\times n\times 6n + 1\times 1\times 6n\times n]$ = $12n^2HW$. With a simple calculation, it is easy to see that if $m=(2/\sqrt{3})n\approx 1.155n$, \textit{\#}MACs/\textit{\#}parameters for \BNReLUConv\ and IBNs are equal. 

Even if a regular \BNReLUConv\ layer has similar \textit{\#}MACs/\textit{\#}parameters as an IBN block, it is well known that IBN blocks achieve significantly higher accuracy~\cite{mobilenetv2}. This is because the IBN blocks have a much higher \textit{expressivity} than the regular convolution layers due to a large number of non-linear units in IBN~\cite{montufar2014number,raghu2017expressive}. Specifically, for our example above (see Fig.~\ref{fig:setup}(a)), the total number of non-linear ReLU units in \BNReLUConv\ = $m$, whereas the IBN has $6n+6n=12n$ ReLU units. Hence, even if $m=1.155n$ (so that both IBN and regular convolution layer have similar \textit{\#}MACs/\textit{\#}parameters), the IBN has more than $10\times$ higher number of non-linear units than \BNReLUConv. This results in better expressivity of IBN-based models. Note that, for AI accelerators, a regular convolution executes much faster than an IBN block due to a significantly better hardware utilization~\cite{mlsys22Codesign,mbdets}, especially if they have similar \textit{\#}MACs/\textit{\#}parameters. Therefore, it is still preferable to have some regular convolution layers in the model.

Based on the above observations, we formulate the task of how much non-linearity can be removed from a network as a novel search problem that chooses between low non-linearity blocks like regular convolutions (e.g., \BNReLUConv) and high non-linearity blocks like IBNs (see Fig.~\ref{fig:setup}(b)). Hence, our search problem is:
\begin{equation}
    \min_{\bm{\theta,\alpha}} \mathcal{L}(y,f(\bm{\theta,\alpha})) + \lambda||\mathbb{I}(\bm{\alpha})||_0,
    \label{eq:search}
\end{equation}
where, $\mathcal{L}$ is the cross-entropy loss, $y$ is the true label for the given classification task, $f$ is the function represented by the \textit{SuperNet} created using the search space, $\bm{\theta}$ are the weight parameters of the SuperNet, and $\bm{\alpha}$ are the parameters that select between low and high non-linearity blocks inside the SuperNet (e.g., \BNReLUConv\ vs. IBNs). The indicator function $\mathbb{I}$ produces a vector whose elements are 1 if a non-linear unit is present, and 0 otherwise; thus, the $l_0$ norm of this indicator function quantifies the number of non-linear units in the model. Finally, $\lambda$ is a hyperparameter that controls the contribution of the second loss. Therefore, the goal of the above problem is to minimize the cross-entropy loss during training and also reduce the total number of non-linear units in the network.

A standard way to solve problem~\eqref{eq:search} can be using any NAS algorithm including differentiable NAS methods like DARTS~\cite{darts}. Towards this, a traditional SuperNet to select between IBN and regular convolution layer would involve putting IBNs and regular convolutions as branches and then an $\alpha$ parameter can select among those options. However, this SuperNet is computationally expensive and requires high GPU memory due to multiple branches. To overcome these issues, we introduce new \textit{Activation Function Restructuring Blocks} (AFRB) that do not require multiple branches (and, hence, are low cost), and enable us to study the non-linearity problem in a very systematic way.

\begin{figure}[tb]
\centering
\includegraphics[width=0.46\textwidth]{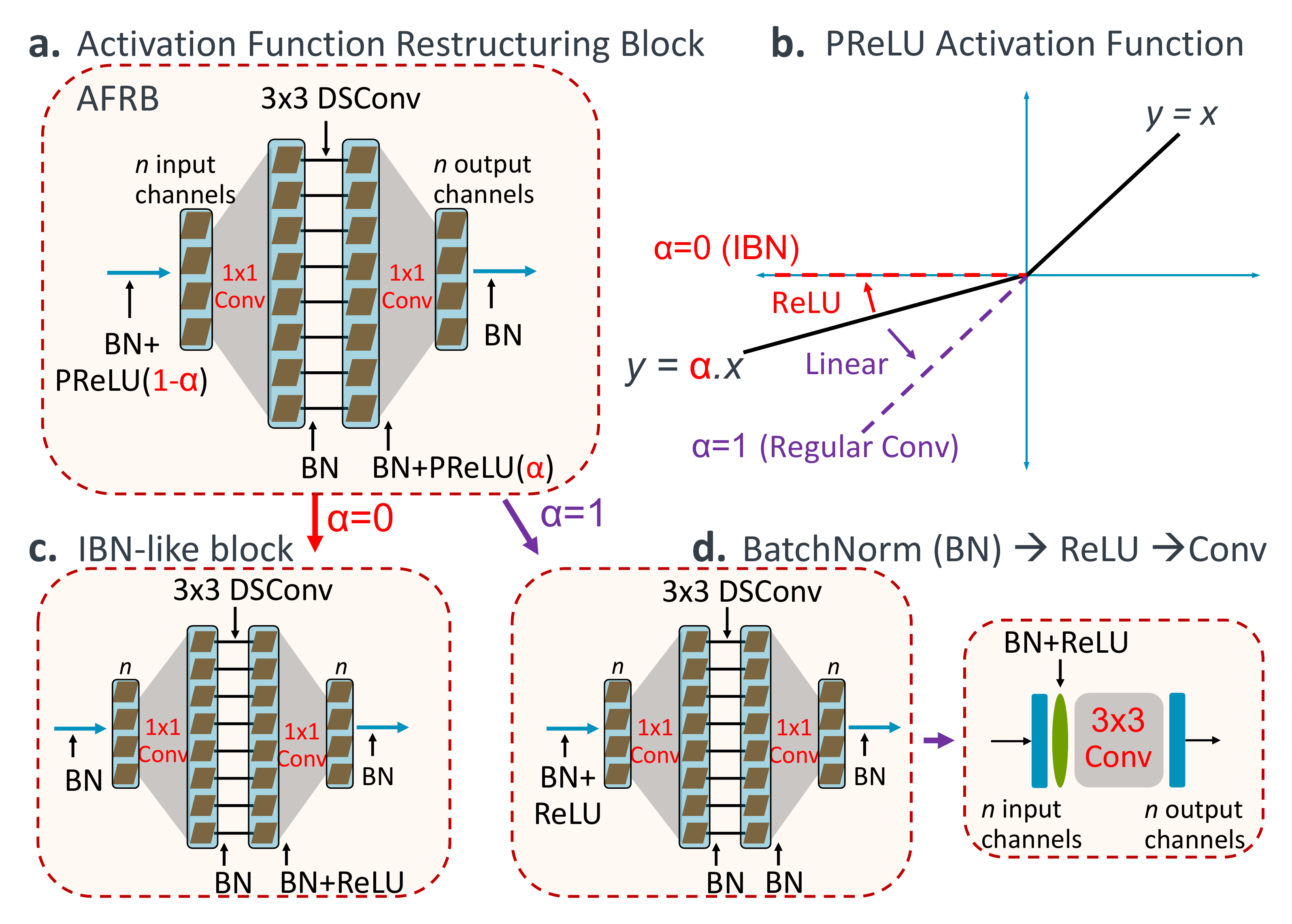}\vspace{-3mm}
	\caption{(a)~Proposed Activation Function Restructuring Block (AFRB) consists of an IBN structure with PReLU activation functions. The first PReLU uses $(1-\alpha)$ and the second one uses $\alpha$ parameter (same $\alpha$ is shared between the two PReLUs). (b)~PReLU activation function: if $\alpha=0$, it becomes ReLU, and if $\alpha=1$, the function becomes linear (i.e., no activation). (c, d)~For $\alpha=0$ ($\alpha=1$), AFRB becomes an IBN-like (a \BNReLUConv) block. Thus, AFRB can restructure an IBN block into a regular convolution by removing non-linear units from the network. This is accomplished using a single scalar trainable $\alpha$ parameter.}
\label{fig:afrb}
\end{figure}

\subsection{New Activation Function Restructuring Blocks}
\label{sec:raneAFRB}
In this section, we propose our Activation Function Restructuring Blocks (AFRB) that automatically restructure IBNs into small, regular $3\times 3$ convolutions. Fig.~\ref{fig:afrb}(a) illustrates the proposed AFRB that consists of an IBN-like structure with two PReLUs. For simplicity, we remove the non-linear activation function after the first $1\times1$ (expansion) layer. Also, the first PReLU appears before the first $1\times1$ convolution and uses $(1-\alpha)$ as its trainable parameter. On the other hand, the second PReLU uses $(\alpha)$ parameter and appears after the depthwise separable convolution (DSConv). Both PReLUs share the same $\alpha$ value.

The PReLU activation function is very interesting because of its trainable $\alpha$ parameter. Specifically, as shown in Fig.~\ref{fig:afrb}(b), if $\alpha=0$, it behaves as a ReLU and if $\alpha=1$, it becomes linear ($y=x$), i.e., no activation. Therefore, if we control the trainable parameter $\alpha$ for the PReLU, we can use it to systematically remove the non-linear units from a network and analyze its impact on accuracy. In other words, using AFRB in our search space allows us to \textit{prune} out the non-linear units from a network in a fully trainable way.

Let us now examine how AFRB enables an explicit restructuring of the blocks. In Fig.~\ref{fig:afrb}(a), if $\alpha=0$, the first PReLU becomes linear and gets removed, while the second PReLU becomes a ReLU that appears after the DSConv layer. Clearly, this resembles an IBN except for the missing ReLU after the expansion layer (see Fig.~\ref{fig:afrb}(c)); we can easily bring back that ReLU once the search phase is over, i.e., during the final training of the searched subnetwork. In contrast, if $\alpha=1$ in Fig.~\ref{fig:afrb}(a), the first PReLU becomes a ReLU and the second PReLU becomes linear (see Fig.~\ref{fig:afrb}(d)). In this case, after the very first ReLU, there are no more non-linear units in the block. That is, the three layers ($1\times1$ expansion layer, DSConv, $1\times1$ projection layer, along with their BatchNorms) are all linear operations. We know from prior art~\cite{sesr, ExpandNets2020}, these linear layers can be \textit{analytically} collapsed into a single regular convolution. Hence, for $\alpha=1$, the AFRB block restructures into a \BNReLUConv\ block. 

Therefore, AFRB is a unique building block that promotes an \textit{explicit trainable restructuring} of entire operations and directly chooses between an IBN block or a \BNReLUConv. Hence, an AFRB-based search space encourages the discovery of completely new kinds of deep networks. We next discuss the direct implications of our block from a hardware perspective.
\begin{figure*}[tb]
\centering
\includegraphics[width=1.0\textwidth]{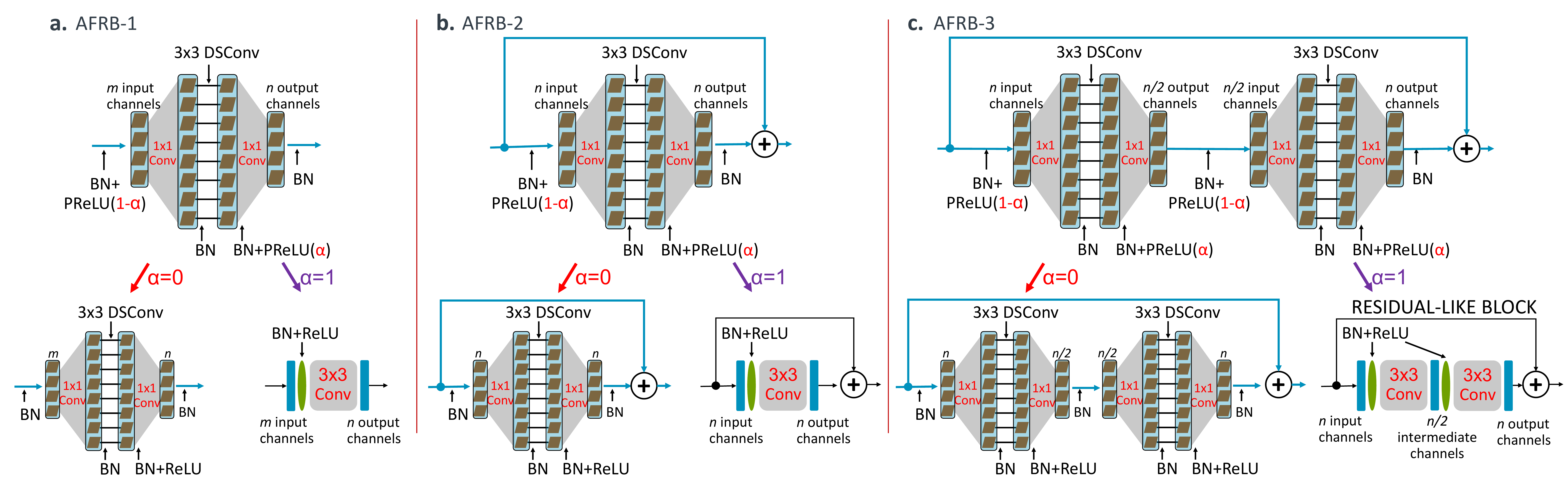}\vspace{-2mm}
	\caption{Proposed search space: (a)~AFRB-1 is used when residual is not present, i.e, if there is a feature map downsampling with stride$>$1 or if \textit{\#}input and \textit{\#}output channels are different. (b)~AFRB-2 is used when a residual can be present (\textit{\#}input channels = \textit{\#}output channels and stride=1). (c)~AFRB-3 can collapse into a residual like block (with half intermediate channels, unlike standard residual blocks). This makes the \textit{\#}MACs/\textit{\#}parameters for AFRB-3 same as AFRB-1 but with additional expressivity (i.e., more non-linear units than AFRB-1).}
\label{fig:ss}
\end{figure*}

\paragraph{Hardware Advantages.} The restructuring of AFRB from IBN to \BNReLUConv\ reduces the computational and memory costs and also improves the hardware utilization on AI accelerators. Specifically, as mentioned earlier, the \textit{\#}MACs for IBN in Fig.~\ref{fig:afrb}(c) = $12n^2HW$, whereas that for \BNReLUConv\ in Fig.~\ref{fig:afrb}(d) = $9n^2HW$. Hence, the restructuring directly results in $25\%$ savings in \textit{\#}MACs/\textit{\#}parameters. Since the regular convolutions do not have utilization issues on AI accelerators, these lower MACs execute at much higher rate on the accelerator, thereby significantly boosting the hardware performance. 

\subsection{Proposed Hardware-Aware Search Space}
\label{sec:raneSS}
We now exploit AFRB to create a novel search space that will be used to generate accelerator hardware-aware models. Fig.~\ref{fig:ss} shows different blocks used to create our SuperNet. As evident, we use three kinds of blocks: (1)~AFRB-1 is used when \textit{\#}input channels are different from \textit{\#}output channels or if there is a stride to downsample feature maps; (2)~AFRB-2 is used when a valid residual skip connection can be added to the block (i.e., \textit{\#}input channels = \textit{\#}output channels and stride = 1); (3)~AFRB-3 can collapse into a residual-like block if $\alpha=1$. Also, AFRB-3 reduces the number of intermediate channels to half the input channels. The idea here is to increase the number of non-linear units while still keeping \textit{\#}MACs/\textit{\#}parameters the same as AFRB-1. A search over the SuperNet containing the above blocks results in accelerator hardware-aware deep networks.

\subsection{Lightweight Search Algorithm}
\label{sec:raneSA}
Let us now explain how to search using our proposed blocks. We first create a SuperNet using AFRBs where each block $i$ (see Fig.~\ref{fig:setup}(b)) has its own $\alpha_i$ parameter. Starting with problem~\eqref{eq:search}, it is easy to see that the trainable $\alpha_i$ parameters make it possible to pick between low non-linearity blocks (\BNReLUConv) and high non-linearity blocks (IBN). The search problem now becomes: 
\begin{equation}
    \min_{\bm{\theta,\alpha}} \mathcal{L}(y,f(\bm{\theta,\alpha})) + \lambda||\mathbb{I}(\bm{\alpha}\in\{0,1\})||_0,
    \label{eq:search1}
\end{equation}
where, the indicator function $\mathbb{I}$ now takes non-zero values only if $\alpha_i=0$ or $\alpha_i=1$ for each block $i$ in the network. In practice, we relax this problem further by trying to make $\alpha_i=1$ where possible and not putting any constraint to make $\alpha_i=0$. That is, instead of searching between linear (no activation, $\alpha_i=1$) or ReLU ($\alpha_i=0$) in problem~\eqref{eq:search1}, we use the following loss to perform a binary search between linear ($\alpha_i=1$) or non-linear ($\alpha_i\neq1$):
\begin{equation}
    \min_{\bm{\theta,\alpha}} \mathcal{L}(y,f(\bm{\theta,\alpha})) + \lambda||\bm{\alpha}-\bm{1}||_2^2,
    \label{eq:search2}
\end{equation}
where, $\bm{1}$ is a vector of all 1's. The above objective function aims to minimize the cross entropy loss while making as many $\alpha_i=1$ as possible. The blocks where $\alpha_i\neq1$ are assumed to be high non-linearity blocks like IBNs. Therefore, minimizing problem~\eqref{eq:search2} can directly restructure some of the IBNs into regular convolutions.

We make a clear distinction between our search phase and final finetuning phases of the initial searched model (in terms of final minor changes to the model architecture). Again, since our search problem only focuses on whether we can remove non-linear units from each block or not, we do not conduct a full-blown NAS in this work. Specifically, number of channels, expansion ratios, number of blocks, etc., are all decided manually when designing the SuperNet. \textit{Our goal is to start with this given SuperNet and make it accelerator-hardware-aware without losing significant accuracy} and \textit{not} to build the most effective search algorithm over traditional factors like number of channels, expansion ratios, number of blocks, etc. To this end, our search will be \textit{semi-automatic}, and we will show how each design decision affects the quality of the model. Nevertheless, we will demonstrate the effectiveness of the proposed search space and our lightweight search algorithm in creating highly efficient deep networks. In the next section, we present this semi-automatic process and show the results on ImageNet.
\begin{figure*}[tb]
\centering
\includegraphics[width=1.0\textwidth]{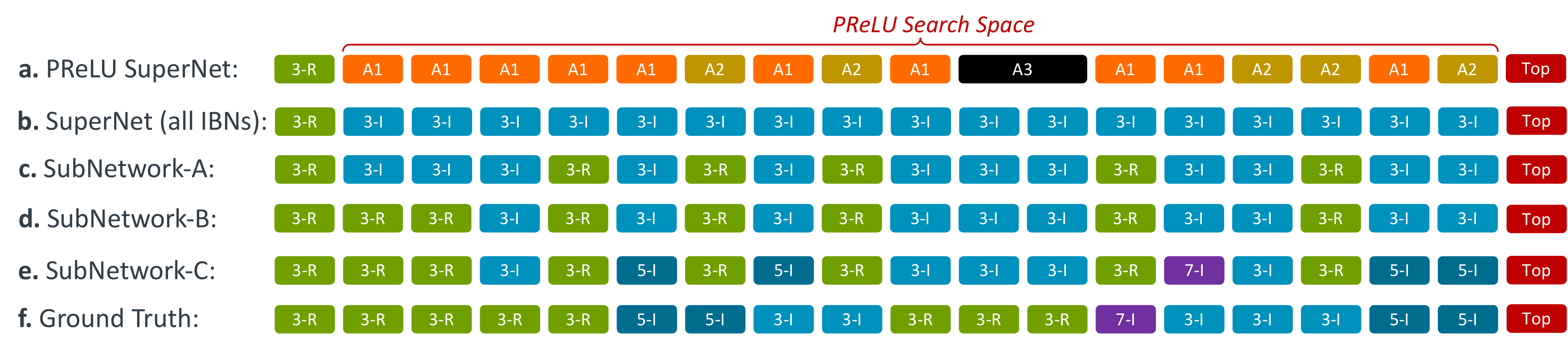}\vspace{-4mm}
	\caption{(a)~\textbf{PReLU SuperNet} shows the SuperNet structure in terms of AFRB blocks. A1, A2, A3 stand for AFRB-1, AFRB-2, and AFRB-3, respectively. The search is conducted over these blocks to see which of them can be restructured into regular convolutions. (b)~\textbf{SuperNet (all IBNs)} consists of 1 regular stem convolution and 17 IBN blocks with $3\times3$ kernel sizes (``3'' in 3-R or 3-I denotes kernel size). (c)~\textbf{SubNetwork-A:} The PReLU-based \textit{accuracy-driven} search reveals that 5 out of 17 blocks can be made regular convolutions without significant accuracy loss. (d)~\textbf{SubNetwork-B:} The first two IBNs are very heavy in MACs due to large feature maps and expansion ratio $e=6$; these blocks are a very natural choice to be collapsed to regular convolutions (on top of SubNetwork-A) to improve hardware performance with some accuracy drop. (e)~\textbf{SubNetwork-C:} We now modify the kernel sizes on some of the IBNs (since the depthwise MACs are still low) to recover the lost accuracy. (f)~\textbf{Ground Truth:} Manual model designed on the SuperNet performs even better than SubNetwork-C (same accuracy and even better hardware performance), thus, highlighting the importance of the proposed search space.}
\label{fig:sr}
\end{figure*}

\begin{table*}[]
\caption{Results on ImageNet (100 epoch training) for the SuperNet and different RAN-e networks (structures shown in Fig.~\ref{fig:sr}). These models are suitable for microcontroller- and mobile-scale AI accelerators. \textcolor{ForestGreen}{Green}/\textcolor{red}{Red} indicate \textcolor{ForestGreen}{improvements} or \textcolor{red}{worsening} of various metrics compared to the original EfficientNet-Lite-B0 (with ReLU6)~\cite{effliteB0}. Normalized throughput is Frames Per Second (FPS) normalized w.r.t. EfficientNet-Lite-B0 estimated for Arm Ethos-U55 (M-class) and Arm Ethos-U65 (A-class) microNPUs.\vspace{-2mm}}
\centering
\scalebox{0.76}{
\begin{tabular}{|l||c|c|c||cc|}
\hline
\multirow{3}{*}{100 epoch training} & \multirow{3}{*}{\begin{tabular}[c]{@{}c@{}}\textit{\#}Parameters\\ (in Millions)\end{tabular}} & \multirow{3}{*}{\begin{tabular}[c]{@{}c@{}}\textit{\#}MACs\\ (in Millions)\end{tabular}} & \multirow{3}{*}{\begin{tabular}[c]{@{}c@{}}Top-1\\ Accuracy\end{tabular}} & \multicolumn{2}{c|}{\begin{tabular}[c]{@{}c@{}}\vspace{-4mm}\\Normalized Throughput (FPS)\end{tabular}}                                                 \\ \cline{5-6} 
                  &                                                                &                                                                &                                                                & \multicolumn{1}{c|}{\begin{tabular}[c]{@{}c@{}}Ethos-U55\\ M-class systems\end{tabular}} & \begin{tabular}[c]{@{}c@{}}Ethos-U65\\ A-class systems\end{tabular} \\ \hline\hline

EfficientNet-Lite-B0~\cite{effliteB0} (original with ReLU6)   & 4.7M  & \textbf{385M}  & $72.81\%$  & \multicolumn{1}{c|}{$1\times$}                                               &   $1\times$                                            \\ \hline
EfficientNet-Lite-B0 (with H-Swish)  & 4.7M  & \textbf{385M}  &  $\bm{73.82\%}$ (\textcolor{ForestGreen}{$\bm{+1.01\%}$}) & \multicolumn{1}{c|}{$1\times$}                                              &    $1\times$                                           \\ \hline\hline
SuperNet (all IBNs)  & 4.7M  & 590M  &  $74.24\%$ (\textcolor{ForestGreen}{$+1.43\%$}) & \multicolumn{1}{c|}{\textcolor{red}{$0.86\times$}}                                           &  \textcolor{red}{$0.87\times$}                                             \\ \hline
SubNetwork-W (reduced width SuperNet, all IBNs)  & \textbf{4.13M}   &  488M & $73.75\%$ (\textcolor{ForestGreen}{$+0.94\%$})  & \multicolumn{1}{c|}{\textcolor{red}{$0.91\times$}}                                              &  \textcolor{red}{$0.90\times$}                                             \\ \hline
RAN-e SubNetwork-A (PReLU search, trained with H-Swish) &  4.6M & 561M  & $73.26\%$ (\textcolor{ForestGreen}{$+0.45\%$})  & \multicolumn{1}{c|}{\textcolor{red}{$0.88\times$}}                                              &    \textcolor{red}{$0.91\times$}                                           \\ \hline
RAN-e SubNetwork-B (SubNetwork-A + first 2 IBNs collapsed) & 4.6M  & 482M  &  $72.63\%$ (\textcolor{red}{$-0.18\%$}) & \multicolumn{1}{c|}{\textcolor{ForestGreen}{$1.10\times$}}                                              &  \textcolor{ForestGreen}{$1.34\times$}                                             \\ \hline
RAN-e SubNetwork-C (SubNetwork-B + new kernel sizes)  & 4.7M  & 488M  & $\bm{73.13\%}$ (\textcolor{ForestGreen}{$\bm{+0.32\%}$})  & \multicolumn{1}{c|}{\textcolor{ForestGreen}{$\bm{1.06\times}$}}                                              &  \textcolor{ForestGreen}{$\bm{1.28\times}$}                                             \\ \hline\hline
RAN-e Ground Truth (manual model within same SuperNet)   & \textbf{4.5M}  & \textbf{433M}  & $\bm{73.04\%}$ (\textcolor{ForestGreen}{$\bm{+0.23\%}$})  & \multicolumn{1}{c|}{\textcolor{ForestGreen}{$\bm{1.16\times}$}}                                              &   \textcolor{ForestGreen}{$\bm{1.49\times}$}                                            \\ \hline
\end{tabular}
}
\label{tab:sr}
\end{table*}

\subsection{RAN-e: ImageNet Evaluation}
\label{sec:ExpRane}
We now exploit our proposed search space to create state-of-the-art networks for the ImageNet image classification task. Fig.~\ref{fig:sr}(a) shows the structure of our SuperNet in terms of the location and types of AFRBs. The complete details of the SuperNet (e.g., strides, channel counts, expansion ratios, etc.) are shown in Table~\ref{tab:supernet} in Appendix~\ref{app:snetFull}. Note that, our SuperNet uses only $3\times3$ kernel sizes for all blocks (see green ``3-R'' blocks for regular convolutions or blue ``3-I'' blocks for IBNs in Fig.~\ref{fig:sr}(b)). As we go through our semi-automatic search process, we will make a number of simple finetuning steps (including kernel sizes) to arrive at the final architecture. More details about the training setup for RAN-e are given in Appendix~\ref{app:raneSetup}.

Table~\ref{tab:sr} shows the \textit{\#}parameters, \textit{\#}MACs, accuracy, and normalized throughput (FPS) of SuperNet and various RAN-e models w.r.t. EfficientNet-Lite-B0~\cite{effliteB0}. The normalized throughput is obtained using performance estimators for Arm Ethos-U55~\cite{u55} (with M-class system configuration) and Arm Ethos-U65~\cite{u65} (with A-class system configuration) microNPUs\footnote{The performance estimator for Ethos-U55 is available at: \url{https://git.mlplatform.org/ml/ethos-u/ethos-u-vela.git/about/}. The performance estimator for Ethos-U65 is proprietary.}. All models in Table~\ref{tab:sr} use Hard-Swish (H-Swish)~\cite{mbv3} activation function unless indicated otherwise. Finally, Table~\ref{tab:sr} results are only for 100 epoch training. We will present 350 epoch training results later.

As evident from Table~\ref{tab:sr}, our SuperNet (assuming all blocks are selected as IBNs, see Fig.~\ref{fig:sr}(b)) achieves $74.24\%$ top-1 accuracy on ImageNet in 100 epochs with 4.7M parameters and 590M MACs. Our SuperNet is slower than EfficientNet-Lite-B0, e.g., $0.87\times$ the FPS of EfficientNet-Lite-B0 on Ethos-U65. This is because the SuperNet incurs a significantly heavier MAC cost (590M vs. 385M) while still only using IBNs in its architecture which present hardware utilization issues for NPUs at some of the layers. Staying within the same IBN-based search space, we further created SubNetwork-W which is simply a reduced width version of the SuperNet (width multiplier = $0.91\times$, see Table~\ref{tab:sr}). Clearly, this model achieves a similar accuracy as EfficientNet-Lite-B0-H-Swish (around $73.8\%$). Again, since this model also uses IBNs only, even with $17\%$ lower MACs than the SuperNet (488M vs. 590M), SubNetwork-W achieves only $3\%$ boost in normalized FPS ($0.9\times$ vs. $0.87\times$) on Ethos-U65. Therefore, hardware utilization issues make it very difficult to improve FPS even if MACs are greatly reduced. Hence, our objective is to make the SuperNet accelerator-hardware-aware using our proposed AFRB blocks and the semi-automatic search algorithm.

\subsubsection{The Initial Search}
We now perform the search, i.e., problem~\eqref{eq:search2} on the PReLU SuperNet shown in Fig.~\ref{fig:sr}(a). The $\lambda$ hyperparameter in problem~\eqref{eq:search2} controls the tradeoff between accuracy and the amount of non-linearity pruning. We performed a search on $\lambda \in (6\times10^{-4}, 2\times10^{-3})$ and found that $\lambda=1\times10^{-3}$ results in minimal accuracy drop ($\sim1\%$) over the SuperNet while collapsing 5 out of 17 IBN blocks into regular convolutions. We call this model SubNetwork-A (see Fig.~\ref{fig:sr}(c)). Recall that, our search only chooses between $\alpha=1$ (linear) and $\alpha\neq1$ (non-linear). The linear case automatically restructures to a regular convolution (see Fig.~\ref{fig:ss}), whereas we set all the high non-linearity $\alpha\neq1$ blocks as IBNs. Since the first activation function is missing in AFRB blocks (see Fig.~\ref{fig:ss}), we bring it back for all $\alpha\neq1$ blocks and retrain the models from scratch once the search phase is complete\footnote{SubNetwork-A (and others) use H-Swish everywhere during retraining and not PReLU. The PReLU is used only during the search phase. In practice, for the automatically restructuring ($\alpha=1$, linear) case, the PReLU search yields $\alpha$ values close to 1 and not exactly 1 when we solve problem~\eqref{eq:search2}, e.g., $\alpha$'s are within a reasonable boundary like $[0.8,1.3]$. When we train the searched SubNetwork-A from scratch, we replace the AFRBs with $\alpha\in [0.8,1.3]$ with regular convolutions.}.  Table~\ref{tab:sr} shows that SubNetwork-A achieves $1\%$ lower accuracy than the all-IBN SuperNet, reduced the \textit{\#}MACs by 29M, and slightly improves the normalized FPS from 0.87 to 0.91 on Ethos-U65. 

Clearly, the performance is not significantly better than the SuperNet. This is because we have only conducted an \textit{accuracy-driven} search and the total number of MACs is still very high compared to the SuperNet (561M vs. 590M). Hence, we next finetune the SubNetwork-A architecture obtained with our initial PReLU search.

\subsubsection{Finetuning Stage 1: Collapse High MAC Blocks}
It is easy to see that our search algorithm chooses to \textit{not} restructure some MAC-heavy blocks, e.g., the first two IBNs in Fig.~\ref{fig:sr}(c). This decision by our PReLU search algorithm makes intuitive sense. Having powerful initial blocks can have a significant impact on accuracy. Since, unlike most hardware-aware NAS works~\cite{proxylessnas, efficientnet}, we have \textit{not} done any MAC- or latency-aware search, the objective in problem~\eqref{eq:search2} attempts to maximize only the accuracy without trying to reduce MACs. This is why it chooses to keep the first two blocks as IBNs even if they are very expensive in MACs. Hence, as our first finetuning stage, we directly restructure the first two IBN blocks -- two of the highest MAC blocks due to large feature map sizes and high expansion ratios ($e=6$) -- into regular convolutions. This results in SubNetwork-B model (see Fig.~\ref{fig:sr}(d)). 

Table~\ref{tab:sr} demonstrates that SubNetwork-B immediately reduces about 80M more MACs over SubNetwork-A and collapsing these two IBN blocks leads to a top-1 accuracy of $72.63\%$ (about $0.6\%$ lower than SubNetwork-A). However, because these newly restructured regular convolutions execute on the AI acclerators without hardware utilization issues, the normalized FPS greatly increases from 0.91 to 1.34 on Ethos-U65 (about $1.47\times$ increase compared to SubNetwork-A). This also highlights the power of our hardware-aware search space: \textit{Even with similar MACs as SubNetwork-W which uses only the traditional IBNs, our model has $1.49\times$ better FPS than SubNetwork-W}.

\subsubsection{Finetuning Stage 2: Change Kernel Sizes}
Inevitably, the accuracy drops when we collapse the first two IBNs in the last finetuning stage. To make up for this lost accuracy, recall that it is very common to use kernel sizes larger than $3\times3$ in modern NAS search spaces (e.g., EfficientNet-Lite also uses $5\times5$ kernel sizes, etc.). Therefore, in this finetuning stage, we increase kernel sizes to $5\times5$ and $7\times7$ for a few IBN blocks. Since the depthwise convolutions in IBNs still incur a low MAC cost, this does not increase the overall MACs significantly. The new updated model is called SubNetwork-C and its structure is shown in Fig.~\ref{fig:sr}(e). As evident from Table~\ref{tab:sr}, SubNetwork-C recovers nearly all of the accuracy to the level of SubNetwork-A ($73.13\%$ vs. $73.26\%$) while still improving normalized FPS from 0.91 to 1.28. SubNetwork-C is the last stage of our semi-automatic search process and clearly demonstrates a systematic way to go from a completely \textit{hardware-unaware} SuperNet to a very \textit{hardware-friendly} SubNetwork. In summary, \textit{compared to EfficientNet-Lite-B0 (original, with ReLU6) trained on our 100 epoch setup, we achieve slightly better accuracy and nearly $1.28\times$ higher FPS}.

\subsubsection{Is this the best we can do?}
So far, the process has been semi-automatic with most of the design decisions being very straightforward after the initial search. We now evaluate if we can do any better with a manual design, within the same search space and SuperNet. To this end, we created a model shown in Fig.~\ref{fig:sr}(f) called the Ground Truth network. Table~\ref{tab:sr} demonstrates that \textit{RAN-e Ground Truth achieves slightly better accuracy than EfficientNet-Lite-B0 (ReLU6) with nearly $1.5\times$ higher FPS which is even better than that for the SubNetwork-C}.

We call this manual model the Ground Truth network because ideally we expect the search algorithm to discover this or a better network on its own. However, since our semi-automatic algorithm produced a sub-optimal network (even if it is $1.28\times$ better than a very strong baseline), it highlights the current limitations of the search process, i.e., lack of a full-blown NAS and no MAC-aware losses. It is possible that with a complete NAS (e.g., channel counts, expansion ratios, kernel sizes, strides, etc.) on the AFRB-based search space, along with explicit MAC-driven losses, the search algorithm may produce an even better network than the RAN-e Ground Truth model. Integrating AFRBs into a full NAS and MAC-constraints is left as a future work.
\begin{table}[]
\caption{ImageNet results. $^\dag$RepVGG and ResNet accuracies are taken directly from~\cite{repvgg} which were trained to 120 epochs. RAN-e accuracies for 100 epoch experiments (Table~\ref{tab:sr}) are higher than ResNet and RepVGG below. $^\ddag$EfficientNet-Lite-B0 as reported by~\cite{effliteB0} using a different training recipe. The remaining RAN-e and EfficientNet-Lite-B0 are trained to 350 epochs on our setup. \vspace{-2mm}}
\centering
\scalebox{0.68}{
\begin{tabular}{|l||c|c|c||cc|}
\hline
\multirow{3}{*}{} & \multirow{3}{*}{\begin{tabular}[c]{@{}c@{}}Params\\\end{tabular}} & \multirow{3}{*}{\begin{tabular}[c]{@{}c@{}}MACs\\\end{tabular}} & \multirow{3}{*}{\begin{tabular}[c]{@{}c@{}}Top-1\\\end{tabular}} & \multicolumn{2}{c|}{\begin{tabular}[c]{@{}c@{}}\vspace{-4mm}\\Normalized FPS\end{tabular}}                                                 \\ \cline{5-6} 
                  &                                                                &                                                                &                                                                & \multicolumn{1}{c|}{\begin{tabular}[c]{@{}c@{}}Ethos-U55\\ M-class \end{tabular}} & \begin{tabular}[c]{@{}c@{}}Ethos-U65\\ A-class \end{tabular} \\ \hline\hline
                  
ResNet-18$^\dag$~\cite{resnet, repvgg}  & 11.7M  & 1.8B  &  $71.2\%$ & \multicolumn{1}{c|}{\textcolor{red}{$0.46\times$}}                                           &  \textcolor{red}{$0.48\times$}                                             \\ \hline
RepVGG-A0$^\dag$~\cite{repvgg}  & 8.3M   &  1.4B & $72.4\%$ & \multicolumn{1}{c|}{\textcolor{red}{$0.72\times$}}                                              &  \textcolor{red}{$0.69\times$}                                             \\ \hline
EffNet-Lite-B0-R6$^\ddag$~\cite{effliteB0}   & 4.7M  & \textbf{385M}  & $75.1\%$  & \multicolumn{1}{c|}{$1\times$}                                               &   $1\times$                                            \\ \hline\hline
EffNet-Lite-B0-R6   & 4.7M  & \textbf{385M}  & $74.4\%$  & \multicolumn{1}{c|}{$1\times$}                                               &   $1\times$                                            \\ \hline 
EffNet-Lite-B0-HS & 4.7M  & \textbf{385M}  &  $\bm{75.6\%}$ & \multicolumn{1}{c|}{$1\times$}                                              &    $1\times$                                           \\ \hline\hline
\textbf{RAN-e-C (Ours)}  & 4.7M  & 488M  & $\bm{74.6\%}$   & \multicolumn{1}{c|}{\textcolor{ForestGreen}{$\bm{1.06\times}$}}                                              &  \textcolor{ForestGreen}{$\bm{1.28\times}$}                                             \\ \hline
\textbf{RAN-e-GT (Ours)}   & \textbf{4.5M}  & \textbf{433M}  & $\bm{74.6\%}$  & \multicolumn{1}{c|}{\textcolor{ForestGreen}{$\bm{1.16\times}$}}                                              &   \textcolor{ForestGreen}{$\bm{1.49\times}$}                                            \\ \hline
\end{tabular}
}
\label{tab:comp}
\end{table} 

\subsubsection{Comparison against reference models}
The original EfficientNet-Lite-B0 (ReLU6) is used only as a reference in the previous sections to show that AFRB-based search space can come up with competitive models. For a fair comparison with our models (that use H-Swish), Table~\ref{tab:sr} also presents the accuracy for EfficientNet-Lite-B0 (H-Swish) trained on our setup. As evident, our SubNetwork-C (RAN-e-C) and Ground Truth (RAN-e-GT) networks are within $1\%$ accuracy of EfficientNet-Lite-B0 (H-Swish) while improving FPS by up to $1.28\times$-$1.5\times$. Note that, EfficientNet-Lite-B0 was obtained using a full-blown NAS that searched over number of blocks, channel counts, expansion ratios, and kernel sizes. Again, because our search is \textit{not} a complete NAS, better networks in AFRB search space could have been potentially obtained if we had also searched over width, depth, expansion ratios, etc. More interestingly, the FPS gain in RAN-e is despite the fact that both SubNetwork-C and Ground Truth networks require more MACs than EfficientNet-Lite-B0. This is perhaps a surprising result because most of the prior art tries to minimize the \textit{\#}MACs/\textit{\#}parameters to obtain efficient models. Therefore, a \textit{hardware-aware search space} can significantly boost performance even with slightly higher MACs.

Next, Table~\ref{tab:comp} shows comparisons against a few existing baselines that also rely on regular convolutions, e.g., ResNets~\cite{resnet} and RepVGG~\cite{repvgg}. We also train our RAN-e networks and EfficientNet-Lite-B0 (ReLU6 and H-Swish) to 350 epochs and report their accuracy. As evident from Table~\ref{tab:comp}, while models like ResNet-18 and RepVGG-A0 use only $3\times3$ convolutions, they result in extremely compute intensive models. Specifically, compared to RAN-e-GT (Ground Truth), ResNet-18 and RepVGG require $4.15\times$ and $3.23\times$ more MACs, respectively. The accuracy for these models reported in Table~\ref{tab:comp} is taken directly from~\cite{repvgg} which trains them to 120 epochs only (without advanced data augmentation techniques like Mixup~\cite{mixup}). We also do not use Mixup in our experiments and our 100 epoch accuracies in Table~\ref{tab:sr} already exceed ResNet-18 and RepVGG accuracies. Furthermore, the normalized FPS on Ethos-U55 and Ethos-U65 clearly demonstrate the superiority of RAN-e-GT compared to ResNet-18 and RepVGG (we achieve up to $3.1\times$ and $2.1\times$ higher FPS, respectively). Note that, the improvements for M-class Ethos-U55 are significant but not as much as the A-class Ethos-U65 because M-class systems are at tiny microcontroller-scale and are limited in memory.

Finally, we have also reproduced EfficientNet-Lite-B0 (ReLU6 (R6) and H-Swish (HS)) accuracies on our setup. Again, while we are $1\%$ away in top-1 accuracy compared to EfficientNet-Lite-B0-HS, we achieve about $1.5\times$ higher FPS. On the other hand, compared to the original EfficientNet-Lite-B0-R6 (trained on our 350 epoch setup), our proposed RAN-e networks achieve about $0.2\%$ higher top-1 accuracy on ImageNet. Of note, with a different training recipe,~\cite{effliteB0} reports an accuracy of $75.1\%$ for EffNet-Lite-B0-R6 (compared to $74.4\%$ for our setup). Therefore, the accuracy for our models may improve even more if the training recipe is optimized further.

\vspace{-3mm}
\paragraph{Other Remarks.} We emphasize that the quality of SuperNet matters in terms of compute costs of different SubNetworks. Specifically, even though both SubNetwork-W and EfficientNet-Lite-B0 (H-Swish) are based on IBNs-only search space and achieve a similar accuracy, SubNetwork-W requires 100M more MACs than EfficientNet-Lite-B0 (see Table~\ref{tab:sr}). Hence, if we had a better, more efficient SuperNet (e.g., in terms of number of blocks, channel counts, expansion ratios, stride locations, etc.), our results could be improved further. This also highlights that (1)~our new hardware-aware search space will likely offer best results when used in conjunction with full NAS, and (2)~perhaps newer, cheap activation functions can also be proposed in future that are specifically designed to work with AFRBs to obtain even higher accuracy (see discussion in Section~\ref{sec:expCon}).

Is it possible to manipulate the non-linearities in deep networks to create accelerator-hardware-aware models? We have demonstrated that an AFRB-based search space enables us to accomplish this goal. The discussion so far completes the proof-of-concept of our novel search space and shows that RAN-explicit is a new direction to achieve hardware-awareness. Next, we propose the implicit restructuring of non-linearities to reduce compute cost of models.

\section{RAN-implicit and Training-Free Scaling}
\label{sec:rani}
Given a good base model, can changing the amount of non-linearity in a network allow us to scale it up or down in a \textit{training-free} way to obtain highly accurate deep networks that satisfy specific \textit{\#}MAC/\textit{\#}parameter constraints? To address this question, we first revisit recent literature~\cite{nnmass} that links topological characteristics of deep networks (i.e., structural properties such as presence of skip connections, etc.) with their gradient propagation and model performance. We also briefly review the literature that studies expressivity of deep neural networks~\cite{montufar2014number,raghu2017expressive, serra2018bounding}.

\subsection{Preliminaries}\label{sec:raniPrelim}
We start by discussing the topological metric and the main theoretical result in~\cite{nnmass}.

\begin{definition}[NN-Mass~\cite{nnmass}]\label{def:nnMass} NN-Mass is defined as the sum (over all $N_b$ blocks) of product of total \textit{\#}input channels ($i_b$) at all layers in a block and Cell-Density ($\rho_b$).
\begin{equation}
	\begin{aligned}
 		\nnMass &= \sum_{b=1}^{N_b} i_b \times \rho_b, \ \ \ \ \ \rho_b = \frac{\text{\textit{\#}Actual Skip Connections}}{\text{Total Possible Skip Connections}}
	\end{aligned}
\label{eq:mass1}
\end{equation}
\end{definition}
For DenseNet-type models\footnote{In~\cite{nnmass}, DenseNet-type networks contain concatenation-type skip connections and the density of skip connections can be varied.}, NN-Mass was shown to be related to the average degree, i.e., average number of connections for each channel in the network -- both short-range, i.e., layer-by-layer connections, and long-range, i.e., skip connections going across multiple layers. Specifically, for a network with width $w$ at all blocks,~\cite{nnmass} proved that the average degree $\hat{k}=w+m/2$. Intuitively,~\cite{nnmass} argues that if we have two networks with similar average connectivity (e.g., same width and NN-Mass), then the amount of information flowing through them is constrained similarly. Therefore, such topological constraints can also have a profound impact on how learning happens in different networks. 

\begin{proposition}[NN-Mass vs. Dynamical Isometry~\cite{nnmass}]
\label{prop1}
Given a deep linear DenseNet-type MLP network, the Layerwise Dynamical Isometry (LDI) is defined as the mean singular value of layerwise Jacobians at initialization. Then, the LDI is bounded as follows:
\begin{equation}
    \sqrt{q\hat{k}}-\sqrt{qw} \leq \mathbb{E}[\sigma] \leq \sqrt{q\hat{k}}+\sqrt{qw},
    \label{eq:prop1}
\end{equation}
where, $\hat{k}=w+m/2$, $q$ is the initialization variance. If $q=1/\hat{k}$, then the mean singular value of layerwise Jacobians (LDI) is bounded near 1, i.e., the gradients flow through the network without amplification or attenuation~\cite{sigprop}.
\end{proposition}

Proposition~\ref{prop1} shows that models with similar width and NN-Mass have similar gradient flow properties and, thus, training convergence (irrespective of their depths, \textit{\#}parameters, and \textit{\#}MACs) since their mean singular value is bounded in a similar way. Therefore, models with similar width and NN-Mass can achieve a similar accuracy even if they have highly different \textit{\#}parameters/\textit{\#}MACs/\textit{\#}layers. Extensive empirical results were presented in~\cite{nnmass} to demonstrate the effectiveness of NN-Mass.

The existing gradient flow-based theory in~\cite{nnmass} shows the relationship between model topology and training convergence but does \textit{not} say anything about expressivity of deep networks. Understanding the expressivity of deep networks is just as important as understanding their gradient properties~\cite{montufar2014number, raghu2017expressive, serra2018bounding}. One way to quantify expressivity of deep networks is to count the number of linear regions that a function represents. These definitions are given below:


\begin{definition}[Linear Regions~\cite{montufar2014number}]
\label{def:linReg}
Given a function $f$, a linear region is a maximal connected set of inputs $x$ where $f$ is linear.
\end{definition}

The number of linear regions can be found by counting the number of unique activation patterns, e.g., how different ReLU units are activating for different inputs~\cite{serra2018bounding}. That is, counting the number of unique activation patterns quantifies how many different linear regions are contained in the function represented by the given deep network. Therefore, this can be directly used as a measure of expressivity of the model. Mont\'{u}far \textit{et al.}~\cite{montufar2014number} provide an aysmptotic lower bound on maximal number of linear regions as follows:
\begin{proposition}[Linear Regions for ReLU Networks~\cite{montufar2014number}]\label{prop:mont}
Given an input $x\in \mathbb{R}^{n_0}$ and a rectifier (ReLU) deep network function $f:\mathbb{R}^{n_0}\rightarrow\mathbb{R}^n$ with $n_0$ input neurons, $L$ layers with $n$ neurons each ($n\geq n_0$), $f$ can compute functions with $\Omega((n/n_0)^{(L-1)n_0}\times n^{n_0})$ linear regions.
\end{proposition}

In the next section, we demonstrate for ResNet- or ConvNext-type networks that NN-Mass is related to the expressivity of deep networks. This is particularly important because while~\cite{nnmass} theoretically analyzes DenseNet-type networks, it neither discusses \textit{why} NN-Mass works for ResNets and other models with residual additions, nor provides any connection with the expressivity of deep networks. 
\begin{figure}[tb]
\centering
\includegraphics[width=0.48\textwidth]{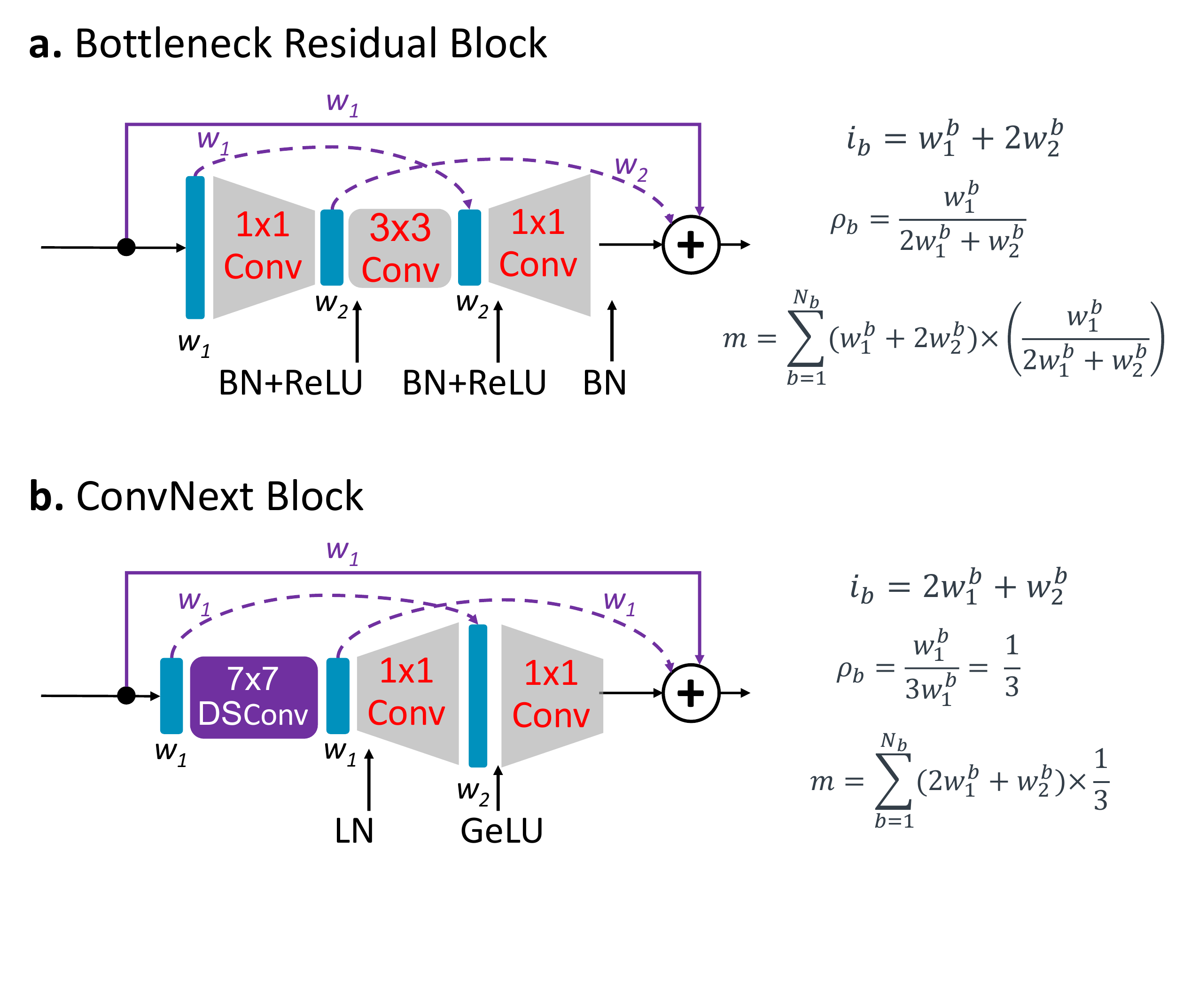}\vspace{-12mm}
	\caption{(a)~NN-Mass calculation for ResNet bottleneck block. (b)~NN-Mass calculation for ConvNext block. In both cases, the solid skip connection is the one actually present in the blocks. The dotted skip connections are shown as \textit{possible} skip connections in the blocks as required by the cell-density ($\rho_b$) definition. Since the skip connections for both blocks involve channel-wise additions, the solid skip connection supplies $w_1$ links, i.e., the skip connection carries information from  all $w_1$ input channels (at first layer); similar ideas apply to possible (dotted) skip connections.}
\label{fig:nnMass}
\end{figure}

\subsection{Expressivity vs. NN-Mass}\label{sec:express}

Bhardwaj \textit{et al.}~\cite{nnmass} empirically derive the NN-Mass expression for ResNet-type networks as shown in Fig.~\ref{fig:nnMass}(a). We also adapt the NN-Mass for ConvNexts (see Fig.~\ref{fig:nnMass}(b)). Note that, for models with a very uniformly repeating structure (e.g., residual blocks or ConvNext blocks), the intermediate width ($w_2$) is generally related to the input channels for the block ($w1$) as $w_2 = e\cdot w_1$, where $e$ is the expansion or shrinking factor and is generally the same for all blocks in the network. Therefore, for this special class of networks, we have the following result:

\begin{proposition}[NN-Mass vs. Linear Regions]\label{prop:nnmassLinReg}
For models with residual additions and a uniform structure (e.g., ResNets/ConvNexts), NN-Mass is proportional to the total number of non-linear units in the network. That is, if the total number of non-linear units in the network = $\mathcal{X}$, then $\mathcal{X}\propto \nnMass$ or $\mathcal{X} = k\times \nnMass$, where $k$ is a constant. Then, upper bound on maximal number of linear regions = $2^{\mathcal{X}} = 2^{k\nnMass}$. 
\end{proposition}

\noindent
\textit{Proof.} Assuming $w_2 = e\cdot w_1$ for all blocks, NN-Mass for ResNets ($m_R$) is given by (see Fig.~\ref{fig:nnMass}(a)):
\begin{equation}
    m_R = \sum_{b=1}^{N_b}(w_1^b(1+2e))\left(\frac{w_1^b}{(2+e)w_1^b}\right) = \left(\frac{1+2e}{2+e}\right)\sum_{b=1}^{N_b}w_1^b
\label{eq:mr}
\end{equation}
Total number of non-linear units for ResNets ($\mathcal{X}_R$) is:
\begin{equation}
    \mathcal{X}_R  = \sum_{b=1}^{N_b}2w_2^b = \sum_{b=1}^{N_b}2e\times w_1^b = 2e\sum_{b=1}^{N_b}w_1^b
\label{eq:xr}
\end{equation}
Therefore, $\mathcal{X}_R\propto m_R$ or $\mathcal{X}_R = k_R\times m_R$, where $k_R = (2e(2+e))/(1+2e)$. A similar calculation for ConvNext class of networks reveals that the total number of non-linear units for ConvNexts ($\mathcal{X}_C$) is also proportional to its NN-Mass ($m_C$), i.e., $\mathcal{X}_C = k_C\times m_C$, where $k_C = (3e/(2+e))$. 

As a result, for models with residual additions and uniform structure, NN-Mass is proportional to the total number of non-linear units in the network. Using~\cite{montufar2014number}, the maximal number of linear regions is bounded by $2^\mathcal{X} = 2^{km}$. Hence, NN-Mass directly impacts the number of linear regions and, thus, the expressivity of this class of deep networks. \qed

Next, we prove one more result on the relationship between NN-Mass and expressivity. Specifically, NN-Mass has known limitations, e.g., depending on the difficulty of the given task, the models need to be deep enough and wide enough for NN-Mass to produce best results in practice\footnote{This was evident in some of the results presented in~\cite{nnmass} (e.g., $R^2$ for relationship between accuracy and NN-Mass increases with increasing width, see Fig.~15 in Appendix H.4 of~\cite{nnmass}). We saw similar patterns in our own experiments as well.}. Proofs in~\cite{nnmass} explicitly assumed conditions on depth while deriving the results, so it is understood that the models need to be deep. However, it is unclear why the width also impacts the effectiveness of NN-Mass. We analyze this relationship between width and NN-Mass below.
\begin{corollary}[NN-Mass matters less when width is low]\label{coro:nnmass}
Suppose we are given a deep network with $n_0$ input neurons, $L$ layers with $n$ neurons each and repeating residual addition skip connections. For such a network, NN-Mass $m = nL$. Then, the function represented by this model $f: \mathbb{R}^{n_0}\rightarrow \mathbb{R}^n$ can compute $\Omega((n/n_0)^{(m-n)\frac{n_0}{n}}\times n^{n_0})$ linear regions. When the width is low (e.g., when $n\rightarrow n_0$), model's expressivity is mostly determined by its width.
\end{corollary}

\noindent
\textit{Proof.} Since all layers have the same width $n$, and because the network consists of repeating residual blocks (similar to Fig.~\ref{fig:nnMass}(a)), the expansion/shrinking factor $e=1$. Substituting this in Eq.~\eqref{eq:mr}, we obtain NN-Mass $m = \sum_{b=1}^{N_b}w_1^b = nL$. For the remaining proof, we simply substitute $L=m/n$ in Proposition~\ref{prop:mont}. Then, it is easy to see that if $n\rightarrow n_0$, $n/n_0 \rightarrow 1$ and, thus, the term with NN-Mass does not contribute to the maximal number of linear regions. Instead, the bound is mostly dictated by the second term that depends only on $n$. This explains why NN-Mass depends on model width: When the width is low, NN-Mass matters less and the expressivity depends more on width. \qed
\begin{figure*}[tb]
\centering
\includegraphics[width=1.0\textwidth]{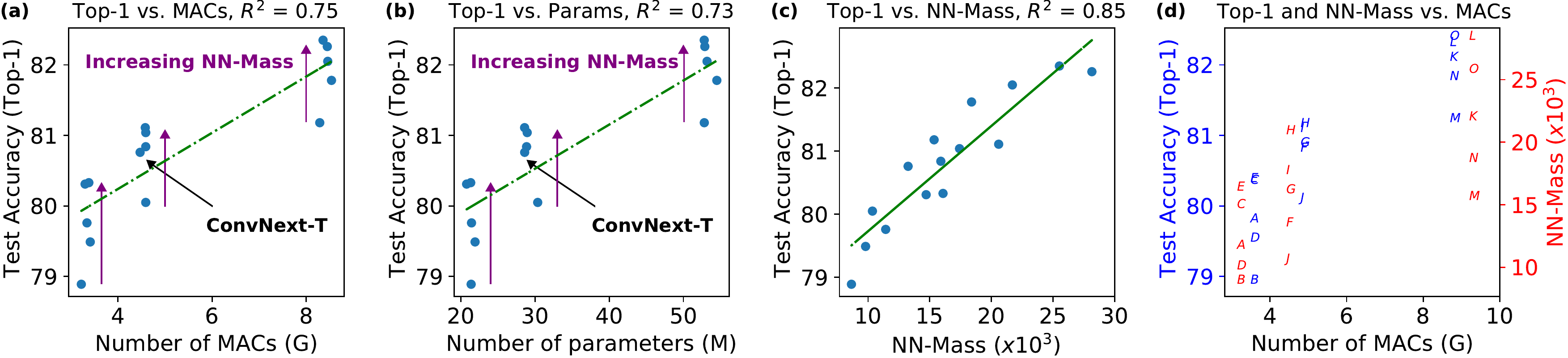}
	\caption{ImageNet accuracy for models scaled up or down from the base ConvNext-Tiny network. All networks are trained for 100 epochs. (a)~Top-1 accuracy vs. \textit{\#}MACs for three MAC budgets: 3.3G, 4.5G, and 8.5G MACs ($R^2=0.75$). (b)~Top-1 accuracy vs. \textit{\#}parameters for three parameter count budgets: 21M, 28M, and 50M ($R^2=0.73$). (c)~Accuracy increases with NN-Mass ($R^2=0.85$). (d)~Each letter denotes a model: blue shows its test accuracy and red shows its NN-Mass. Clearly, higher NN-Mass results in higher test accuracy for all models except O and L. Accuracy saturates as NN-Mass becomes high for a given constraint (O and L have less than $0.1\%$ accuracy gap).}
\label{fig:raniNNMass}
\end{figure*}

Therefore, model topology not only impacts the gradient properties as proved in~\cite{nnmass}, it also directly affects the expressive power of deep neural networks. Since computing metrics like NN-Mass does not require any training or even a single forward or backward pass, they can serve as excellent training-free methods to search for high quality models.

\subsection{Training-Free Scaling with NN-Mass}\label{sec:trainFree}
We now exploit NN-Mass for training-free model scaling. Specifically, given a base model $\mathcal{M}$ and a few different hardware constraints $\bm{\mathcal{H}}=\{H_1,H_2,\ldots,H_p\}$ (in terms of \textit{\#}MACs/\textit{\#}parameters), the problem is to scale the model up or down in a training-free fashion to find high quality models for all constraints. Once an appropriate model is found using this training-free \textit{search}, it is trained to obtain the final model. Note that, all other conditions from the last section must hold true for model $\mathcal{M}$: It should have a large depth and width, and must consist of a uniform block structure that repeats throughout the network.

Since computing NN-Mass does not require training, we scale the base model $\mathcal{M}$ as follows: We scale the depth and width of $\mathcal{M}$ using a set of width and depth multipliers, e.g., $W \in ({w}_{min}, {w}_{max})$ and $D\in ({d}_{min},{d}_{max})$. For each $w_m\in W$ and $d_m\in D$, we compute the $\{$\textit{\#}MACs, \textit{\#}parameters, NN-Mass$\}$. Then, for the models that satisfy the given hardware budget constraint $H_i\in \bm{\mathcal{H}}$, we pick the model with highest NN-Mass and train it. Based on the theory discussed in Section~\ref{sec:express}, we expect this model to have superior gradient flow properties and expressive power than other models with similar \textit{\#}MACs/\textit{\#}parameters. Therfore, model $M_i$ for hardware budget $H_i$ is simply the network that maximizes NN-Mass for that hardware budget. Since \textit{\#}MACs/\textit{\#}parameters/NN-Mass do not require training, this process is completely training free.

Note that, scaling the base model by $w_m$ and $d_m$ \textit{implicitly} restructures the non-linear activation functions because it changes NN-Mass which in turn changes the total number of non-linear units in the network (see Proposition~\ref{prop:nnmassLinReg}). Therefore, we call our networks RAN-implicit (RAN-i). We next present detailed experimental results on ImageNet to show the effectiveness of our training-free scaling.

\subsection{RAN-i: ImageNet Evaluation}\label{sec:expRani}
We start with the base model ConvNext-Tiny~\cite{convnext} and scale it to three hardware budgets: (1)~$H_1$: 3.3B MACs and 21M parameters, (2)~$H_2$: 4.5B MACs and 28M parameters, i.e., the same hardware budget as ConvNext-Tiny, and (3)~$H_3$: 8.5B MACs and 50M parameters. To sample models that satisfy the above hardware budgets, we use width multipliers $W\in (0.25,1.6)$ and depth multipliers $D\in (0.6,2.56)$. In total, we sampled 800 different models using the above width and depth multipliers. For each network, we then compute $\{$\textit{\#}MACs, \textit{\#}parameters, NN-Mass$\}$. We found total 15 networks that satisfied $\bm{\mathcal{H}}=\{H_1,H_2,H_3\}$ budgets defined above (i.e., 5 models for each of $H_1, H_2, H_3$). As explained in Section~\ref{sec:trainFree}, for each budget, the higher NN-Mass models are expected to achieve the higher accuracy due to better gradient properties and expressivity. To verify this, we trained all 15 networks on ImageNet for 100 epochs to evaluate if this is indeed the case. Detailed training setup is given in Appendix~\ref{app:raniSetup}. 

We show the top-1 accuracy and its relationship with \textit{\#}MACs/\textit{\#}parameters/NN-Mass for all 15 networks in Fig.~\ref{fig:raniNNMass}. Note that, all models belong to the same ConvNext family of networks and the only difference is that their widths and depths are scaled up or down from ConvNext-Tiny (ConvNext-T) network. Yet, Fig.~\ref{fig:raniNNMass}(a,b) show that for exactly the same hardware budget, there can be a significant difference in accuracy. We found that for all three hardware budgets, models with increasing NN-Mass result in higher accuracy (see Fig.~\ref{fig:raniNNMass}(c)). Specifically, $R^2=0.85$ for NN-Mass is higher than that for \textit{\#}MACs ($R^2=0.75$) and \textit{\#}parameters ($R^2=0.73$). 

Fig.~\ref{fig:raniNNMass}(d) also shows the exact ranking of top-1 accuracy and NN-Mass vs. \textit{\#}MACs. Here, each letter denotes a model, blue color shows its accuracy, and red color shows its NN-Mass. Going from top to bottom for each hardware budget, we can see that the ranking for accuracy and NN-Mass is the same for almost all cases\footnote{For instance, going from top to bottom for 3.3B MACs in Fig.~\ref{fig:raniNNMass}(d), we see the ranking of networks as E-C-A-D-B for both NN-Mass and accuracy, thus showing that higher NN-Mass results in higher accuracy.}. Note that, for each hardware budget, accuracy eventually saturates and, hence, increasing NN-Mass stops improving the models. This is visible from models O and L in Fig.~\ref{fig:raniNNMass}(d).  This also indicates that NN-Mass should have some lower and upper bounds for it to work optimally. While we have derived some of these conditions in Corollary~\ref{coro:nnmass}, more theoretical analysis can certainly improve our understanding of NN-Mass. Nevertheless, our results clearly demonstrate that NN-Mass significantly cuts down the search space of possible depths and widths by providing us the most promising candidates in a completely training-free manner.
\begin{figure}[tb]
\centering
\includegraphics[width=0.48\textwidth]{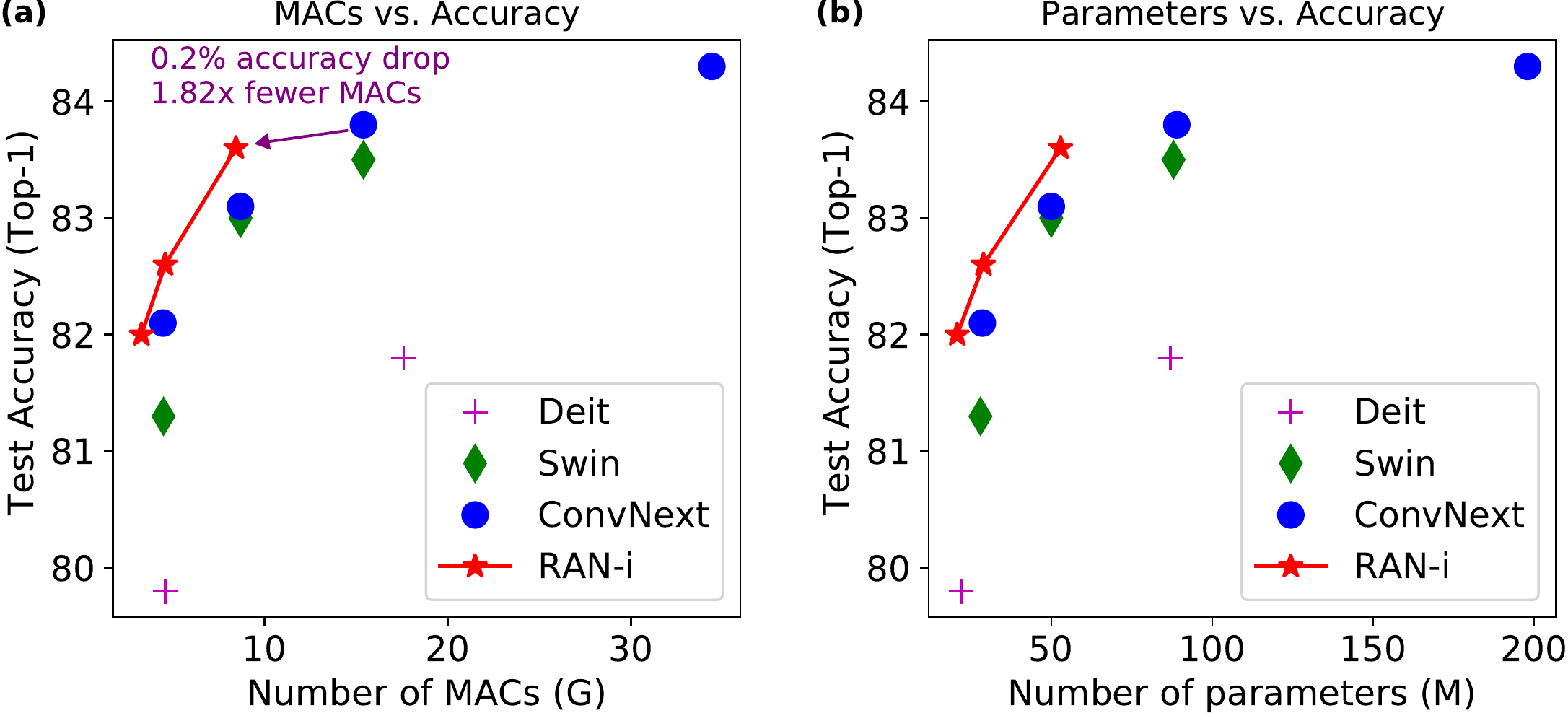}\vspace{-1mm}
	\caption{ImageNet results: RAN-i networks (i.e., highest NN-Mass models for various hardware budgets) achieve state-of-the-art accuracy and establish a new pareto frontier over existing networks. Models are trained for 300 epochs. (a)~MACs vs. Top-1 Accuracy: RAN-i can achieve $83.6\%$ top-1 accuracy which is only $0.2\%$ lower than ConvNext-B~\cite{convnext} while requiring $1.82\times$ fewer MACs. (b)~Parameter count vs. Top-1 Accuracy: RAN-i leads to significant savings in number of parameters as well.}
\label{fig:raniFinal}
\end{figure}

Finally, we pick the highest NN-Mass networks for each hardware budget and train them to 300 epochs on ImageNet. We call these highest NN-Mass networks as RAN-i; more architecture details are given in Appendix~\ref{app:raniArch}. The results are shown in Fig.~\ref{fig:raniFinal}. Clearly, RAN-i establishes a new state-of-the-art on ImageNet as it beats the Pareto frontier of ConvNexts. Specifically, our 8.45B MACs RAN-i network achieves only $0.2\%$ lower top-1 accuracy than ConvNext-B that requires 15.4B MACs. This results in up to $1.82\times$ fewer MACs with nearly the same accuracy. Significant improvements are also obtained in \textit{\#}parameters. Moreover, for a concrete FPS evaluation, we deploy our 4.59B MACs RAN-i network and 8.7B MACs ConvNext-S on a single core Arm Neoverse-based datacenter CPU. We found that RAN-i achieves nearly $40\%$ ($1.38\times$) higher FPS than ConvNext-S with about $0.5\%$ lower accuracy. Therefore, NN-Mass is a very inexpensive method to push the Pareto frontier and to scale models in a training-free way to various hardware constraints once a good base model is known.

\subsection{Object Detection with NN-Mass Scaling}
To further illustrate the utility of network scaling with NN-Mass, we compare the performance of RAN-i models against that of ConvNext models when deployed as object detection backbones. We design a simple two-stage object detector and show that it performs better when it uses our scaled models than when it uses ConvNexts.

Two-stage object detectors, such as the popular Faster-RCNN and Mask-RCNN, are common because of their ease of use and competitive detection accuracies \cite{fasterrcnn, maskrcnn}. A two-stage detector funnels the input image through a \textit{backbone} network (RAN-i or ConvNext, in this case) and then sends the extracted features to a \textit{Region Proposal Network (RPN)}, which proposes \textit{Regions of Interest (RoIs)} around objects to be classified by the \textit{head}.

Our detector architecture is largely the same as that of Faster-RCNN, but we make the following changes to slim it down: (1) In the RPN, we only use PyTorch's design of a single $3\times3$ convolutional layer, followed by two $1\times1$ convolutions \cite{finetuningtorchvision}. We restrict the model to produce 512 RoIs instead of 800 RoIs. (2) In the head, we only use two fully connected layers with 512 neurons each before the output.

For an apples-to-apples comparison, we compare the performance of our object detector  in two cases: (1) the backbone is RAN-i, and (2) the backbone is ConvNext. As shown in Table~\ref{tab:objdet_results}, when RAN-i backbones are used, object detectors run up to $1.33\times$ faster (measured on an Arm Neoverse-based datacenter CPU) than when ConvNext equivalents are used, while MACs are reduced by $1.83\times$ and parameters by $1.49\times$. On the COCO dataset\cite{coco}, accuracy of our RAN-i backbone model either exceeds or is similar to that of the ConvNext backbone model. It is also competitive with ResNet50-FPN-Faster-RCNN's 36\% mAP \cite{torchcontributors}, despite requiring $1.6\times$ fewer MACs. ResNet50-FPN-Faster-RCNN is a Faster-RCNN model with a ResNet-50 backbone and a Feature Pyramid Network (FPN), intended to handle objects at different sizes~\cite{fpns}. We do not use a FPN in our model. Further, our architecture facilitates easy training, achieving 34.7\% mAP in just 26 epochs (15 hours wall-clock time, see Appendix~\ref{sec:appendix_objdet_training} for training details). Therefore, RAN-i can be used to significantly reduce object detection compute costs without affecting accuracy.
\begin{table}[]
\caption{Object Detection Results. On the COCO dataset, object detectors backboned by NN-Mass scaled RAN-i models (see Appendix~\ref{app:raniArch}) achieve competitive accuracies with significantly less computation requirement than when backboned by ConvNexts.}
\scalebox{0.8}{
\begin{tabular}{|l|c|c|c||c|}
\hline
Backbone   & \multicolumn{1}{c|}{Params} & \multicolumn{1}{c|}{MACs} & \multicolumn{1}{c||}{mAP} & \multicolumn{1}{c|}{FPS Improvement} \\ \hline\hline
ConvNext-S & 94.9M                       & 150.6B                    & $34.4\%$                   &        \multirow{2}{*}{1.33$\times$}               \\ \cline{1-4}  
\textbf{RAN-i-S (Ours)}     & \textbf{63.6M}                       & \textbf{82.5B}                     & $\bm{34.7\%}$                   & 
                            \\ \hline\hline
ConvNext-B & 149.9M                      & 219.4B                    & $\bm{35.0\%}$                   &  \multirow{2}{*}{1.21$\times$}               \\ \cline{1-4}
\textbf{RAN-i-B (Ours)}     & \textbf{93.2M}                       & \textbf{123.4B}                    & $34.9\%$                   & 
                                    \\ \hline
\end{tabular}
}
\label{tab:objdet_results}
\end{table}

The benefits of our object detection experiments are further emphasized by the architecture's ease of use and versatility. By incorporating NN-Mass scalable backbones, an existing detector can be easily modified to achieve substantial computation improvements. The rapid trainability of the design (no layer-freezing required) also facilitates training with fewer resources. The backbone can be replaced by any other feature extractor, and the RPN and head are further modularized. Users of this design can thus strike a comfortable balance between accuracy and computational requirements in a relatively inexpensive manner.

\section{Block and Activation Function Co-Design}\label{sec:expCon}
\begin{figure}[tb]
\centering
\includegraphics[width=0.48\textwidth]{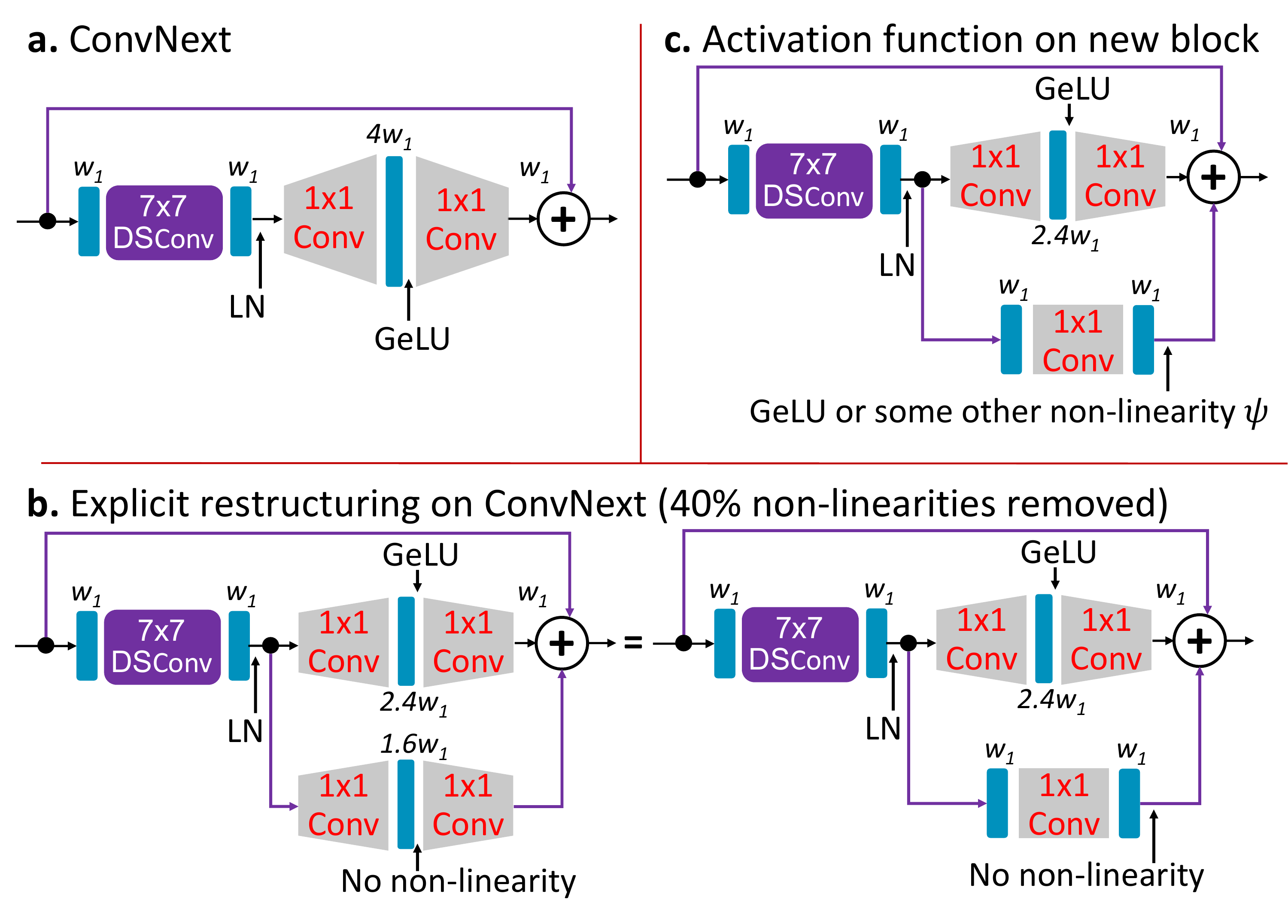}\vspace{-3mm}
	\caption{Block and Activation Function Co-Design: (a)~Typical ConvNext block consists of an inverted bottleneck with $4\times$ expansion ratio (see $4w_1$ GeLUs between the $1\times1$ layers). (b)~If $40\%$ non-linear units are removed, the inverted bottleneck can be analytically separated into two branches. Since the lower branch does not have any non-linearity, it restructures into a single $1\times1$ convolution with $w_1$ input and output channels, thereby saving \textit{\#}MACs/\textit{\#}parameters over the base model. (c)~Can we regain some accuracy by re-introducing cheap non-linearity on second branch?}
\label{fig:expConv}
\end{figure}

So far, we have implicitly restructured the amount on non-linearity in ConvNext. Is it possible to directly manipulate the non-linear units in ConvNext to explicitly restructure it into a different network architecture? If yes, can we co-design novel activation functions that make up for lost expressivity due to restructuring? We note that the ConvNext block also contains an inverted bottleneck kind of structure with a $1\times1$ convolution expanding the $w_1$ input channels by a factor of 4, followed by $4w_1$ non-linear GeLU units, and then a $1\times1$ convolution projecting back to $w_1$ channels (see Fig.~\ref{fig:expConv}(a)). Total \textit{\#}MACs can be significantly reduced if all or at least some of those GeLUs can be linearized. This can be similar to RAN-e that fully removes non-linear activations from expanded channels. However, we found that the accuracy drops significantly in ConvNexts if we  linearize all GeLU units in a block. 

To this end, we focus on the following task: We first remove $40\%$ GeLUs from each block of ConvNext-T network. As shown in Fig.~\ref{fig:expConv}(b), the $40\%$ linearized channels can then be \textit{analytically} separated out as a secondary branch. That is, the primary (upper) branch has $60\%$ channels ($0.6\times4w_1 = 2.4w_1$) with GeLUs, and the secondary (lower) branch has $40\%$ channels ($0.4\times4w_1 = 1.6w_1$) with no non-linear activation function. Then, similar to RAN-e, the lower branch explicitly restructures into a single $1\times1$ convolution. This process reduces the number of MACs from $(H\times W\times w_1\times 4w_1)\times2=8HWw_1^2$ in ConvNext-T to $(H\times W\times w_1\times 2.4w_1)\times2+(H\times W\times w_1^2)=5.8HWw_1^2$, i.e., $27.5\%$ fewer MACs compared to the initial block.

Inevitably, removing non-linear units would result in some loss of accuracy. It is natural to ask if there is anything we can do on the lower branch to recover the lost accuracy. For instance, can we use a GeLU or some other activation function $\psi$ on the lower branch? Note that, the inverted bottleneck structure seems to be a common theme in most of the state-of-the-art models and results in a significant improvement in accuracy. The main defining characteristic of the inverted bottleneck is that the non-linearity is applied in \textit{higher dimensions}, e.g., after expanding the initial number of channels using a $1\times1$ convolution. We hypothesize that this ``applying non-linearity in higher dimensions'' is responsible for high accuracy achieved by most of the networks, e.g., EfficientNets~\cite{efficientnet}, ConvNext~\cite{convnext}, Swin-Transformers~\cite{swin}, etc. Therefore, we ask the following question w.r.t. the lower branch in our restructured block: \textit{Is there an inexpensive way we can operate higher dimensions without increasing computational costs?}

The above question has been very well-studied in the machine learning community. Specifically, the \textit{kernel trick}~\cite{bishop} used in Support Vector Machines~\cite{svm} can project low-dimensional inputs into high-dimensional spaces without ever leaving the original low-dimensional space. Towards this, we consider the non-linear activation function $\psi$ for the lower branch as the exponential function: 
\begin{equation}
    \psi(\bm{x}, \bm{\beta}) = e^{\langle\bm{x},\bm{\beta}\rangle} = \sum_{n=0}^{\infty}\frac{{\langle\bm{x},\bm{\beta}\rangle}^n}{n!},
\label{eq:exp}
\end{equation}
where, $\bm{x}$ is the input data patch and $\bm{\beta}$ is the learnable weight for the $1\times1$ convolution on the lower branch. Clearly, similar to all kernel tricks, the exponential kernel implicitly operates in an infinite-dimensional space without ever explicitly computing the sum in Eq.~\eqref{eq:exp}. Note that, by designing an activation function to make up for the lost expressivity during our explicit restructuring, we are attempting to \textit{co-design a restructurable block with a novel activation function}. We next evaluate whether this co-design can help us achieve a higher accuracy.

Table~\ref{tab:expConv} demonstrates the results for explicit restructuring of ConvNext-T network. Here, all models are trained on ImageNet for 100 epochs. As evident, if we remove the lower branch completely, this results in a $40\%$ channel pruned version of ConvNext-T. This network loses about $1.3\%$ accuracy over the baseline. Next, we evaluate two networks: (1)~A $27.5\%$ channel pruned version of ConvNext-T which is about $0.9\%$ below the baseline, and (2)~A $40\%$ non-linearity restructured network as shown in Fig.~\ref{fig:expConv}(b), right (we call this as Model A). Both of these networks achieve a similar accuracy and incur exactly the same \textit{\#}MACs and \textit{\#}parameters. Next, we train Model B which appends a GeLU activation function at the end of lower branch (see Fig.~\ref{fig:expConv}(c)). Model B achieves only $0.1\%$ higher accuracy than Model A. Finally, we train Model C using the exponential activation function. Even though Model C has exactly the same \textit{\#}MACs and \textit{\#}parameters as Models A, B, and the $27.5\%$ channel-pruned ConvNext-T, it achieves nearly $0.3\%$-$0.4\%$ higher accuracy. This supports our hypothesis that operating in higher dimensions can still be beneficial in deep networks even if it is done using kernel tricks.

While we achieve accuracy improvement with our proposed activation function, there are clear limitations: introducing an exponential in the network makes it highly unstable. Specifically, we observed that during training, it can often lead to NaN loss. However, when the model does train, we get better convergence and accuracy than no-activation and GeLU cases. This is a key limitation of our kernel trick. Therefore, more stable activation functions that implicitly operate in high dimensions should be designed in future.   
\begin{table}[]
\caption{Block and Activation Function Co-Design for ConvNext-Tiny. Models are trained for 100 epochs on ImageNet.}\vspace{-2mm}
\centering
\scalebox{0.74}{
\begin{tabular}{|l||c|c|c|}
\hline
\multirow{2}{*}{100 epoch training} & \multirow{2}{*}{Params} & \multirow{2}{*}{MACs} & \multirow{2}{*}{Top-1} \\
                  &                    &                    &                    \\ \hline\hline
ConvNext-T                 &    28.6M                &    4.47B                &       $80.2\%$             \\ \hline
ConvNext-T ($40\%$ pruned)               &            18.2M        &         2.8B           &     $78.9\%$               \\ \hline
ConvNext-T ($27.5\%$ pruned)                 &        21.5M            &        3.32B            &   $79.3\%$                 \\ \hline
Model A ($40\%$ restructured, Fig.~\ref{fig:expConv}(b) right)                &     21.5M               &       3.32B              &            $79.3\%$        \\ \hline
Model B (Model A + [$\psi=$ GeLU], Fig.~\ref{fig:expConv}(c))                &      21.5M              &               3.32B      &              $79.4\%$      \\ \hline
Model C (Model A + [$\psi=$ EXP], Fig.~\ref{fig:expConv}(c))                &          21.5M          &               3.32B     &                 $\bm{79.7\%}$   \\ \hline
\end{tabular}
}
\label{tab:expConv}
\end{table}

\section{Outstanding Problems}
\label{sec:disc}
So far, we have demonstrated that explicit and implicit restructuring of non-linear activation functions is valuable for deep learning. Based on our insights, we now discuss the following open problems in this new domain:
\begin{itemize}
    \item \textbf{Manipulating Non-Linearity as a Key NAS Goal:} In the AI accelerator age, high utilization yet high accuracy building blocks are a prerequisite for hardware-aware networks. However, existing NAS does not use a search space containing restructurable blocks like AFRBs. Therefore, future NAS research should exploit non-linearity manipulation as a key objective.
    \item \textbf{Non-Linearity and Model Architecture Co-Design:} More generally (i.e., beyond NAS), a new research direction is to co-design novel restructurable blocks along with ways to induce the non-linearity elsewhere in the network. We have attempted to do this in our Section~\ref{sec:expCon}, e.g., we restructured a known block with a more powerful, theoretically-grounded activation function. However, since our exponential function has clear stability issues, more research is needed in this area to potentially discover completely new building blocks that are friendly to AI accelerators.
    \item \textbf{Better Theory and Generic Topological Metrics:} Our  NN-Mass~\cite{nnmass} based method allows us to scale base models to various hardware constraints in a training-free manner and achieves state-of-the-art accuracy on ImageNet. There are still limitations which can be targeted in future work: (\textit{i})~Currently NN-Mass works for large networks. There must be specific bounds on depth and width to obtain optimal results with NN-Mass (see Section~\ref{sec:expRani}). While Corollary~\ref{coro:nnmass} is a step in this direction, more theory is needed to better understand NN-Mass.  (\textit{ii})~Also, NN-Mass works only for uniform structures and does not work for irregular strides and full NAS search spaces. Thus, better topological metrics are required for generic NAS as well.
    \item \textbf{More Theory for Explicit Restructuring:} We need better theoretical grounding for explicit restructuring of activation functions. For instance, how does the dual objective in problem~\eqref{eq:search2} change the deep learning optimization landscape? Can we build better optimizers that specifically work well for AFRBs? Improvements in this space can directly improve the overall accuracy without changing the computational costs.  
\end{itemize}

The research directions above can significantly impact efficient deep learning, particularly for AI accelerators. 

\section{Related Work}
\label{sec:rel}
\paragraph{Linear overparameterization in deep networks. }
The benefit of linear overparameterization in accelerating the training of deep neural networks and improving accuracy has received considerable attention in recent works~\cite{AroraICML2018,GCNICML2019,ACNetICCV2019, ExpandNets2020, DOConv2022, repvgg}. Several of these previous works propose overparameterizing a convolutional layer during training by using a series of linear convolutional layers. More recently, RepVGG~\cite{repvgg} demonstrates the importance of linear residual connections in parallel branches of a neural network during training, which can be folded during inference to boost the accuracy of single-branch networks. In contrast to these prior works on model overparameterization, RAN-e seeks to identify where in the network the non-linear activation functions can be removed. This results in a sequence of linear convolution layers that can be collapsed into a single, small, regular convolution layer. Overall, our approach produces networks that use a mix of IBNs and regular convolutions, and achieve significantly higher accuracy at lower computational cost than RepVGG~\cite{repvgg}.

\paragraph{Non-linearity manipulation.} Concurrent with our work, Fu  \textit{et al.}~\cite{DepthShrinker} proposed DepthShrinker, which combines irregular blocks into dense operations to create hardware-efficient, compact neural networks. DepthShrinker also proposes to replace low-utilization blocks with regular convolutions by pruning non-linear units. Despite the synergies between our work and DepthShrinker, there are significant differences and advantages of our work:
\begin{enumerate}
  \item \textbf{More generality:} RANs are much more general than just non-linearity pruning: We propose a hardware-aware search space for future NAS methods. We further propose training-free model scaling with theoretically grounded non-linearity manipulation, as well as a co-design between blocks and activation functions which could also be useful for networks where complete removal of non-linear units may not be possible.
  \item \textbf{Fully differentiable restructuring:} Our RAN-e is a fully differentiable restructuring algorithm. In contrast, DepthShrinker~\cite{DepthShrinker} relies on approximate techniques like Straight-Through Estimators (STE)~\cite{ste} which can be unstable under certain conditions~\cite{yin2019understanding}.
  \item \textbf{No self-distillation:} DepthShrinker exploits methods like self-distillation to regain the accuracy drop. Distillation-based techniques are known to result in significant accuracy improvements~\cite{donna}. We do not use any distillation-based methods to improve accuracy. 
  \item \textbf{Better accuracy:} RAN-e achieves significantly higher accuracy than DepthShrinker. In particular, on ImageNet, RAN-e-C achieves a $2.1\%$ higher accuracy under comparable MACs compared to DepthShrinker's MBV2-1.4-DS-D model (e.g., 488M MACs RAN-e-C achieves $74.6\%$ accuracy  vs. $72.5\%$ accuracy for 484M MACs DepthShrinker; no self-distillation used in either network). Additionally, for similar MACs, RAN-e-GT trained \textit{without} self-distillation significantly outperforms DepthShrinker trained \textit{with} self-distillation by $\bm{4.47\%}$ in top-1 accuracy (e.g., $74.6\%$ for 433M MACs RAN-e-GT vs. $70.13\%$ for DepthShrinker's 415M MACs MBV2-1.4-DS-F).
  \item \textbf{SotA ImageNet results on multiple scales:} Finally, our techniques result in state-of-the-art results on ImageNet at multiple scales, ranging from micro-NPUs to datacenter CPUs. In contrast, DepthShrinker does not cover such a broad range of ImageNet networks.
\end{enumerate}   

\paragraph{DNN compression techniques. }
Numerous research efforts have been devoted in recent years to compressing neural networks for increasing hardware efficiency in accelerators via filter pruning~\cite{LEGRCVPR2020, LFPCCVPR2020, LotteryTicketICLR2019}, layer pruning~\cite{OptimalBrainSurgeonNeurIPS2017, Elkerdawy_2020_ACCV, XuLayerPruning2020}, quantization~\cite{LSQICLR2020, BannerNeurIPS2019, FBRevisitingQuantICLR2020, FBQuantNoiseICLR2021}, and low-rank matrix factorization~\cite{Yin_2021_CVPR, Tai2015, Wen2017}. Nonetheless, because these model compression techniques are orthogonal to our hardware-aware block search paradigm, they can be combined with our search space to improve hardware efficiency even further.

\paragraph{NAS for improving hardware efficiency. }
Recent research on automated efficient DNN design has been able to take advantage of significant advances in neural architecture search (NAS) to select from a variety of hardware-efficient convolutional blocks, layer widths, depths, connectivities, and per-layer quantization bitwidths during training while building a network architecture~\cite{proxylessnas, micronets_mlsys2021, EfficientNetV2ICML2021,FastModelFamiliesCVPR2021, mbdets, fbnetv2, UNASCVPR2020, MixedPrecisionDNNsICLR2020, MCUNetNeurIPS2020}. For example, MobileDet's NAS search space included IBN as well as other hardware-aware convolutions like fused and tucker convolutions~\cite{mbdets}. While it is possible to naively construct a NAS search space from parallel branches of IBN and hardware-friendly regular convolutions, our work proposes using non-linearity manipulation to choose between IBN and hardware-friendly convolutional blocks from the same underlying weight-shared block. This, unlike previous works, will enable the search process between different convolutional blocks to take advantage of weight-sharing NAS. Manipulating non-linearity can essentially be added as another search dimension during NAS. We leave this exploration for future work.

\section{Conclusion}
\label{sec:conc}
In this paper, we have proposed the new RAN paradigm that manipulates the amount of non-linearity in networks to improve their hardware-efficiency. Specifically, we have proposed RAN-explicit (RAN-e) and RAN-implicit (RAN-i) techniques for hardware-aware search spaces and training-free model scaling, respectively. For certain classes of networks, we have also theoretically proved the link between model expressivity as defined by the amount of non-linearity and its topological properties. With extensive experiments, we have demonstrated that our networks achieve state-of-the-art results on ImageNet at different scales and for various types of hardware ranging from micro-NPUs to datacenter CPUs. Our proposed RAN-e achieves a similar accuracy as EfficientNet-Lite-B0 while improving FPS by up to $1.5\times$ on Arm micro-NPUs. Moreover, with a similar or better accuracy, our RAN-i networks demonstrate nearly $2\times$ reduction in \textit{\#}MACs and about $40\%$ improvement in FPS on Arm Neoverse-based datacenter CPUs compared to ConvNexts. When used as backbones in object detection, RAN-i achieve a similar or higher mAP over ConvNexts with $33\%$ higher FPS on datacenter CPUs. Finally, we have also discussed a new research direction of model architecture-activation function co-design. 

Overall, we have demonstrated several useful scenarios where manipulating non-linear activation functions in deep networks directly results in significant hardware-awareness and efficiency. For future work, we have outlined several outstanding research problems in this new area of restructurable deep networks.

{\small
\bibliographystyle{ieee_fullname}
\bibliography{egbib}
}

\appendix

\section{RAN-e: SuperNet Details}\label{app:snetFull}
The SuperNet architecture details are given in Table~\ref{tab:supernet}. The top architecture for our SuperNet is different from the typical top convolutions used in networks like EfficientNet. The detailed structure of our top is given in Table~\ref{tab:top}.
\begin{table}[h]
\centering
\caption{Detailed architecture for the RAN-e SuperNet. $H_i$ and $W_i$ are height and width of input feature maps, respectively. $\{e, s, C_o\}$ refer to expansion ratio, stride, and output channels at the given stage, respectively. All kernel sizes in SuperNet are $3\times3$.}
\scalebox{0.9}{
\begin{tabular}{|c|c|c|c|c|c|}
\hline
Stage & Operator & $H_i\times W_i$ & $e$ & $s$ & $C_o$ \\ \hline\hline
1    &    Conv $3\times3$      &      $224\times224$           & $-$  & $2$  &    $16$   \\ \hline\hline
2    &    AFRB-1      &      $112\times112$           & $6$  & $1$  &    $32$   \\ \hline
3    &    AFRB-1      &      $112\times112$           & $6$  & $2$  &    $48$   \\ \hline
4    &    AFRB-1      &      $56\times56$           & $6$  & $2$  &    $64$   \\ \hline
5    &    AFRB-1      &      $28\times28$           & $6$  & $1$  &    $80$   \\ \hline
6    &    AFRB-1      &      $28\times28$           & $6$  & $2$  &    $80$   \\ \hline
7    &    AFRB-2      &      $14\times14$           & $6$  & $1$  &    $80$   \\ \hline
8    &    AFRB-1      &      $14\times14$           & $4$  & $1$  &    $96$   \\ \hline
9    &    AFRB-2      &      $14\times14$           & $4$  & $1$  &    $96$   \\ \hline
10    &    AFRB-1      &      $14\times14$           & $6$  & $1$  &    $128$   \\ \hline
11    &    AFRB-3      &      $14\times14$           & $6$  & $1$  &    $128$   \\ \hline
12    &    AFRB-1      &      $14\times14$           & $6$  & $2$  &    $160$   \\ \hline
13    &    AFRB-1      &      $7\times7$           & $4$  & $1$  &    $176$   \\ \hline
14    &    AFRB-2      &      $7\times7$           & $4$  & $1$  &    $176$   \\ \hline
15    &    AFRB-2      &      $7\times7$           & $4$  & $1$  &    $176$   \\ \hline
16    &    AFRB-1      &      $7\times7$           & $6$  & $1$  &    $224$   \\ \hline
17    &    AFRB-2      &      $7\times7$           & $6$  & $1$  &    $224$   \\ \hline\hline
$-$ & Top & $7\times7$ & $-$ & $-$ & $1000$ \\ \hline
\end{tabular}
}
\label{tab:supernet}
\end{table}
\begin{table}[h]
\centering
\caption{Top convolution architecture for RAN-e networks.}
\scalebox{1.0}{
\begin{tabular}{|c|c|c|c|c|}
\hline
Stage & Operator & $H_i\times W_i$ & $C_i$ & $C_o$ \\ \hline
1      &    Conv $1\times1$      &  $7\times7$               &  224     &  1344     \\ \hline
2      &    DSConv $7\times7$      &    $7\times7$             &    1344   &    1344   \\ \hline
3      &    Average Pool      &     $1\times1$            & 1344      &     1344  \\ \hline
4      &    Conv $1\times1$      &  $1\times1$               &  1344     &  1000     \\ \hline
\end{tabular}
}
\label{tab:top}
\end{table}

\section{RAN-e: Training Details}\label{app:raneSetup}
We train all RAN-e networks and SuperNet on ImageNet using Autoaugment data augmentation and label smoothing with value 0.1. We also use RMSprop optimizer with an initial learning rate 0.005 which follows a cosine annealing decay schedule after an initial warmup of 5 epochs, batch size 768, decay 0.9, momentum 0.9, epsilon 0.001, and weight decay 5e-6. We do not use Mixup data augmentation~\cite{mixup} in our experiments. We implement our PReLU search as well as finetuning experiments in Tensorflow. All models are trained on 8 NVIDIA V100 GPUs.

\section{RAN-i: Training Details}\label{app:raniSetup}
Our training setup for RAN-i networks is nearly identical to that used in the ConvNext paper and its public code~\cite{convnext}. The only difference is that we reduced the batch size to 80 for our networks in order to fit within the GPU memory. The batch size was lowered to 80 for both the initial 15 networks sampled using NN-Mass in Fig.~\ref{fig:raniNNMass}, and the final models trained in Fig.~\ref{fig:raniFinal}. For the initial 15 networks in Fig.~\ref{fig:raniNNMass}, we used drop path = 0.1 for all networks. In the next section, we provide more details for the final RAN-i networks that were trained to 300 epochs.

\section{Final RAN-i Architecture Details}\label{app:raniArch}
Table~\ref{tab:raniFinal} more details on the final RAN-i networks. The base width and depth configurations are the same as that for ConvNext-Tiny network. For example, the first group in ConvNext-Tiny has 3 blocks with 96 input and output channels at each block, followed by 3 blocks with 192 channels each, and so on. To obtain RAN-i networks, these base widths and depths are multiplied by width multiplier $w_m$ and depth multiplier $d_m$, respectively. As evident, the resulting network configurations satisfy the hardware constraints like $\{$3.3B, 4.5B, 8.5B$\}$ MACs. The RAN-i networks in Table~\ref{tab:raniFinal} are the highest NN-Mass models for the aforementioned hardware constraints.
\begin{table}[h]
\centering
\caption{Detailed configurations for RAN-i networks. These are the final networks that were trained to 300 epochs in Fig.~\ref{fig:raniFinal}.}
\scalebox{0.76}{
\begin{tabular}{|l||ccc|}
\hline
                                               & \multicolumn{1}{c|}{RAN-i-T (Tiny)} & \multicolumn{1}{c|}{RAN-i-S (Small)} & RAN-i-B (Base) \\ \hline\hline
\begin{tabular}[c]{@{}l@{}}Base Width\\ Config\end{tabular} & \multicolumn{3}{c|}{[96,192,384,768]}                              \\ \hline
\begin{tabular}[c]{@{}l@{}}Base Depth\\ Config\end{tabular} & \multicolumn{3}{c|}{[3,3,9,3]}                              \\ \hline\hline
$w_m$                                             & \multicolumn{1}{c|}{0.666}  & \multicolumn{1}{c|}{0.789}  &  0.9105 \\ \hline
$d_m$                                             & \multicolumn{1}{c|}{1.65}  & \multicolumn{1}{c|}{1.65}  &  2.30 \\ \hline
\begin{tabular}[c]{@{}l@{}}New Width\\ Config\end{tabular} & \multicolumn{1}{c|}{[64,128,256,511]}  & \multicolumn{1}{c|}{[76,151,303,606]}  &  [87,175,350,699] \\ \hline
\begin{tabular}[c]{@{}l@{}}New Depth\\ Config\end{tabular} & \multicolumn{1}{c|}{[5,5,15,5]}  & \multicolumn{1}{c|}{[5,5,15,5]}  & [7,7,21,7]  \\ \hline
\textit{\#}Parameters                                              & \multicolumn{1}{c|}{20.76M}  & \multicolumn{1}{c|}{28.93M}  & 52.89M  \\ \hline
\textit{\#}MACs                                              & \multicolumn{1}{c|}{3.3B}  & \multicolumn{1}{c|}{4.59B}  & 8.45B  \\ \hline
Drop Path                                             & \multicolumn{1}{c|}{0.1}  & \multicolumn{1}{c|}{0.2}  & 0.4  \\ \hline
Top-1                                            & \multicolumn{1}{c|}{$82.03\%$}  & \multicolumn{1}{c|}{$82.63\%$}  & $83.61\%$ \\ \hline
\end{tabular}
}
\label{tab:raniFinal}
\end{table}

\section{Object detection: Training Details} \label{sec:appendix_objdet_training}
Our training setup is nearly identical to that used by Facebook to train their ResNet50-FPN-Faster-RCNN model \cite{torchcontributors}. We do not freeze any layers in our detectors, and start training the RPN and head against ImageNet-pretrained backbones. As with Facebook's procedure, minimal data augmentation is performed (resize image to 800x800, add 50\% probability of horizontal image flips). We use Facebook's modified COCO loss~\cite{coco} with stochastic gradient descent (learning rate controlled by a LR Scheduler, with momentum of 0.9 and weight decay of 0.0001). Training on the Microsoft COCO 2017 dataset lasts for 26 epochs, using roughly 15 hours of wall-clock time on 8 NVIDIA V100 GPUs, each with a batch size of 2.

\end{document}